\RequirePackage[svgnames]{xcolor}

\documentclass[11pt,letterpaper]{mystyle}
\usepackage[all]{hypcap}
\usepackage[svgnames]{xcolor}
\usepackage[comma,authoryear,compress]{natbib}
\usepackage{hyperref}
\usepackage{amsmath}
\usepackage{etoolbox}

\definecolor{darkblue}{rgb}{0, 0, 0.5}
\hypersetup{
    colorlinks = true,
    citecolor = {darkblue},
    linkcolor = {darkblue},
    urlcolor = {darkblue},
}

\usepackage{wrapfig}
\usepackage{algorithm}
\usepackage{algorithmicx}
\usepackage{algpseudocode}
\usepackage{microtype}
\usepackage{booktabs}
\usepackage{float}
\usepackage{comment}
\usepackage{amssymb}
\usepackage{mathtools}
\usepackage{amsthm}
\usepackage{mathrsfs}
\usepackage{nicefrac}
\usepackage{dsfont}
\usepackage{enumitem}
\usepackage{subcaption}
\usepackage{cleveref}

\usepackage[utf8]{inputenc}
\usepackage[T1]{fontenc}
\usepackage{url}
\usepackage{amsfonts}
\usepackage{graphicx}
\usepackage{fdsymbol}
\usepackage{lipsum}
\usepackage{stackengine}
\usepackage[font=small,labelfont=bf]{caption}
\usepackage{color}
\usepackage{adjustbox}
\usepackage{rotating}
\usepackage{makecell}
\usepackage{multirow}

\definecolor{BlueBG}{rgb}{0,0.46,0.71}

\definecolor{blanchedalmond}{rgb}{1.0, 0.92, 0.8}
\definecolor{carmine}{rgb}{0.59, 0.0, 0.09}
\definecolor{lightblue}{rgb}{0.22,0.45,0.70}%

\renewcommand{\mathbf}{\boldsymbol}

\makeatletter
\def\Ddots{\mathinner{\mkern1mu\raise\p@
\vbox{\kern7\p@\hbox{.}}\mkern2mu
\raise4\p@\hbox{.}\mkern2mu\raise7\p@\hbox{.}\mkern1mu}}
\makeatother

\definecolor{amaranth}{rgb}{0.9, 0.17, 0.31}
\definecolor{antiquebrass}{rgb}{0.8, 0.58, 0.46}
\definecolor{antiquefuchsia}{rgb}{0.57, 0.36, 0.51}
\definecolor{chromeyellow}{rgb}{0.31, 0.47, 0.26}

\usepackage{pifont}
\usepackage{soul}
\usepackage{authblk}
\usepackage{bbm}
\usepackage{optidef}

\usepackage{mathpazo}
\usepackage{xspace}
\usepackage{xcolor}
\usepackage{tcolorbox}
\usepackage{tikz}
\usetikzlibrary{positioning, arrows.meta}

\usepackage{arydshln}
\usepackage{siunitx}
\usepackage{listings}
\usepackage{fvextra}
\tcbuselibrary{skins,breakable}
\usepackage{fontawesome5}
\usepackage{lineno}

\newcommand{\tg}[1]{\textcolor{green!60!black}{\scriptsize #1}}
\newcommand{\tr}[1]{\textcolor{red}{\scriptsize #1}}

\title{Models Can Model, But Can't Bind: Structured Grounding in Text-to-Optimization}

\runningtitle{Models Can Model, But Can't Bind: Structured Grounding in Text-to-Optimization}

\author[1]{Zhiqi Gao$^\ast$}
\author[1]{Albert Ge$^\ast$}
\author[2]{Alexander Berenbeim}
\author[2]{Nathaniel D. Bastian}
\author[1]{Frederic Sala}

\affil[1]{University of Wisconsin-Madison}
\affil[2]{United States Military Academy}

\begin{document}

\begin{abstract}
{\centering\section*{Abstract}}

\emph{Text-to-optimization} requires two separable capabilities: \emph{modeling}---choosing the right optimization structure---and \emph{binding}---grounding every coefficient, index, and parameter in the concrete problem data.
We study this via Text2Opt-Bench, a scalable benchmark of solver-verified optimization problems spanning 12 categories, from textbook linear programs to stochastic and multi-objective formulations with up to thousands of variables.
Across 10+ models, we find that accuracy collapses as instance data grows, even when the formulation itself is simple.
We call this the \emph{effective binding limit}.
We study it with a family of techniques, \textbf{BIND}, that externalize numeric data to structured files so the model binds data programmatically rather than transcribing from the prompt.
When using an oracle for externalizing data, we recover between 12 and 27 accuracy points, confirming binding as a key---but recoverable---failure mode. 
In a deployable setting without oracle access, we validate our hypothesis by finetuning a model exclusively on binding and show that it outperforms end-to-end SFT and RL across three structurally distinct optimization categories, with a 1.5B binding specialist alone matching a 7B end-to-end baseline.

\vspace{1em}
{\centering
\faGithub\ \url{https://github.com/SprocketLab/Text2Opt-Bench} \\
\faDatabase\ \url{https://huggingface.co/datasets/ZhiqiGao/Text2Opt-Bench} \\
}

\end{abstract}

\maketitle
\correspondingauthor{Zhiqi Gao: zhiqi@cs.wisc.edu \quad $^\ast$Equal contribution. }

\section{Introduction}

Operations research (OR) is central to industrial decision-making in logistics, energy, and supply chains. Solving OR tasks from natural language with LLMs (performing text-to-optimization) requires two distinct abilities: \textbf{(1) modeling}, i.e., selecting the correct optimization model and structure, and \textbf{(2) binding}, i.e., grounding variables, constraints, coefficients, and other problem parameters to the given data. The first capability requires \emph{reasoning} skills, an area where models have recently made significant progress. The second, however, remains challenging to achieve. We argue that current text-to-optimization systems are primarily bottlenecked by binding rather than modeling.

To test this hypothesis, we turn to benchmarks that measure text-to-optimization capabilities. Existing benchmarks~\citep{pmlr-v220-ramamonjison23a, mostajabdaveh2025evaluatingllmreasoningoperations, wang2024optibench, Huang_2025} address textbook problem scale: small, deterministic, single-objective programs in which every constraint is explicitly stated. Real-world OR involves uncertainty, competing objectives, and domain knowledge that is used to induce constraints. These features are absent from existing benchmarks.

We address these challenges via  \textbf{Text2Opt-Bench}, a scalable benchmark of verified optimization problems spanning \textbf{12 problem categories} covering linear programs (LP), mixed-integer linear programs (MILP), mixed-integer quadratic programs (MIQP), and nonlinear formulations---including stochastic programs with chance constraints, multi-objective formulations with competing cost and emissions targets, and problems requiring domain-specific constraint derivation (Ohm's law, Erlang-C queuing). Our benchmark is built via a \emph{forward-engineering} pipeline: we first construct a solver-verified optimization problem, then generate a natural language description grounded in the problem's underlying scenario parameters. This decouples linguistic generation from mathematical structure, ensuring that each problem instance is feasible by construction and that evaluation failures can be unambiguously attributed to the model rather than to benchmark artifacts.

\begin{figure}[t]
    \centering
    \includegraphics[width=\textwidth]{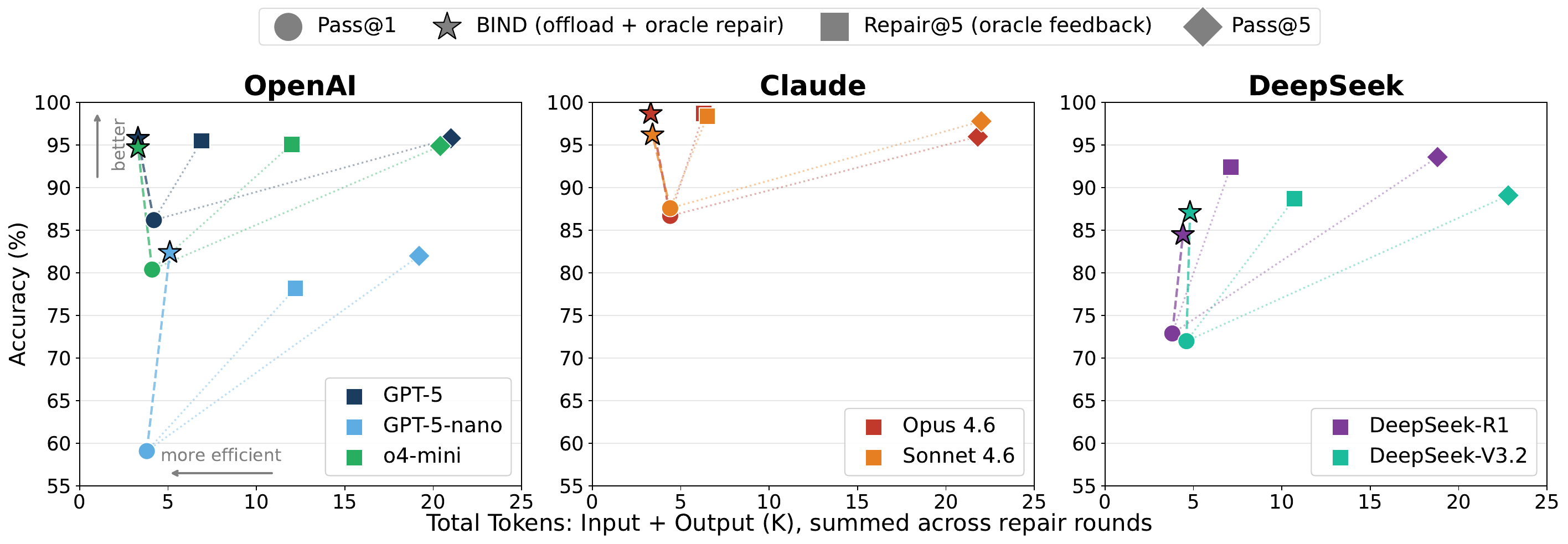}
    \caption{Solution accuracy vs.\ total token cost across three model families on 550 template problems. BIND ($\le$3 oracle-guided refinement rounds) matches oracle Repair@5 and approaches pass@5 at far lower token cost.}
    \label{fig:bind_pareto}
\end{figure}

Using this benchmark, we evaluate 10+ models from OpenAI, Claude, Deepseek, Llama, and Qwen families and report three main findings:

\textbf{(1) For frontier models, binding is the dominant bottleneck on most categories.} For example, GPT-5-Nano's accuracy drops from 72\% to 11\%  as instance data grows, even when the formulation is unchanged. Closed-source frontier models are closely matched at 86--88\% overall, while reasoning models (o4-mini, DeepSeek-R1) fail to surpass standard models, suggesting these do not address binding failures. The same accuracy cliff appears on non-OR RULER retrieval tasks (\S\ref{sec:binding_bottleneck}). 

\textbf{(2) Binding-aware techniques can resolve most binding failures.} We introduce \textbf{BIND}, a family of binding-aware techniques. The data-offload portion externalizes numeric data to structured files so the model binds programmatically rather than transcribing from the prompt. Given oracle-extracted data, a single pass recovers +12 to +27pp on binding-limited categories for \emph{every} model; with $\le$3 rounds of oracle-guided refinement BIND reaches near-ceiling accuracy (GPT-5 95.8\%, GPT-5-Nano 82.4\%). The deployable variant supplies the data via a trained binding specialist that outperforms end-to-end training.

\textbf{(3) Training binding-specific models is an effective recipe across categories we study.} We turn from inference-only approaches to training. Surprisingly, we find that supervised finetuning (SFT) outperforms reinforcement learning (RL) at 7B scale. This is consistent with binding as the bottleneck: SFT provides dense supervision of coefficient transcription while RL's sparse reward struggles to distinguish between a wrong formulation and a wrong parameter. Motivated by this observation, we show that training a 7B binding specialist outperforms end-to-end SFT across three structurally distinct categories: 58.1\% vs. 51.2\% (resource allocation, paired McNemar p=0.002), 100\% vs. 96\% (job-shop scheduling), and 96\% vs. 88\% (transportation).

\noindent In summary, our primary contributions are (1) Text2Opt-Bench, a scalable, solver-verified benchmark of 12 problem categories (LP/MILP/MIQP/nonlinear, up to 1,000+ variables); (2) a binding bottleneck analysis showing that instance binding is the dominant failure mode on most categories, confirmed via RULER retrieval experiments; (3) BIND, a family of binding-aware techniques: an oracle-fed data offload that diagnoses recoverable accuracy, and a trained binding specialist that deploys the same idea without an oracle; and (4) a demonstration that decomposing training by binding yields stronger and more parameter-efficient models than end-to-end SFT or RL.
\section{Related Work}
\label{sec:related_work}

We briefly detail relevant related work.

\textbf{Text-to-Optimization.} There is ongoing work to develop benchmarks and methods for solving optimization problems from natural language. On the benchmark side,
NL4Opt~\citep{pmlr-v220-ramamonjison23a} treats optimization as entity extraction on small LPs. OptiBench~\citep{wang2024optibench}, ORLM~\citep{Huang_2025}, MAMO~\citep{huang2025llmsmathematicalmodelingbridging}, and OptMATH~\citep{lu2025optmathscalablebidirectionaldata} offer solver-verified instances but at textbook problem scale. 
More recent efforts (OPT-Engine~\citep{chen2026optenginebenchmarkinglimitsllms}, ProOPF~\citep{shen2026proopfbenchmarkingimprovingllms}, ConstraintBench~\citep{tso2026constraintbenchbenchmarkingllmconstraint}, ORQA~\citep{mostajabdaveh2025evaluatingllmreasoningoperations}, NLMOptimizer~\citep{berenbeim2025nlmoptimizer}) expand problem types and scale. Table~\ref{tab:benchmark_comparison} compares these benchmarks. Our benchmark, Text2Opt-Bench, offers controllable difficulty, scalability up to 1,000+ variables, and industrially-motivated formulations.

On the methods side, OptiMUS~\citep{ahmaditeshnizi2024optimusscalableoptimizationmodeling} and Chain-of-Experts~\citep{xiao2024chainofexperts} use modular decomposition; LLMOPT~\citep{jiang2025llmoptlearningdefinesolve} learns to define problems end-to-end. 
OR-LLM-Agent~\citep{zhang2025orllmagentautomatingmodelingsolving} decomposes tasks into modeling, coding, and debugging. For a survey, see~\citet{xiao2025surveyoptimizationmodelingmeets}.

\begin{table}[t]
\caption{Comparison with existing OR benchmarks.}
\label{tab:benchmark_comparison}
\begin{center}
\small
\begin{tabular}{lccccc}
\toprule
\textbf{Benchmark} & \textbf{Problems} & \textbf{Verified} & \textbf{Max Vars} & \textbf{Types} & \textbf{Adv.\ Form.} \\
\midrule
NL4Opt & 1,101 & \texttimes & ${\sim}$5 & LP & \texttimes \\
OptiBench & 605 & $\checkmark$ & ${\sim}$50 & Mixed & \texttimes \\
ORLM & 100 & $\checkmark$ & ${\sim}$10 & LP/MILP/NLP & \texttimes \\
MAMO & 1,209 & $\checkmark$ & ${\sim}$50 & LP/MILP/ODE & \texttimes \\
OPT-Engine & 1,810 & $\checkmark$ & ${\sim}$40 & LP/MIP & \texttimes \\
\midrule
\textbf{Ours} & \textbf{scalable} & $\checkmark$ & \textbf{1,000+} & \textbf{LP/MILP/MIQP/NLP} & $\checkmark$ \\
\bottomrule
\end{tabular}
\end{center}
\end{table}


\textbf{Synthetic Data Generation.}
Verifiable synthetic data has proven valuable for reasoning~\citep{liu2025synlogicsynthesizingverifiablereasoning, goldie2025syntheticdatageneration, seegmiller2025flamesimprovingllmmath}; our forward-engineering pipeline differs from back-translation approaches (e.g., OptMATH) by jointly generating descriptions and OR structures from simulated world states. 

\textbf{Data Externalization and Programmatic Access}. A growing body of work offloads context from the prompt to external environments that the model accesses programmatically. PAL \citep{gao2023palprogramaidedlanguagemodels} and Program of Thoughts \citep{chen2023programthoughtspromptingdisentangling} generate code rather than performing computation in-context; Recursive Language Models \citep{zhang2026recursivelanguagemodels} generalize this by treating the entire prompt as an external environment the model can recursively query. These approaches address computational or context-length limitations. BIND targets a different bottleneck — faithful transcription of numerical data — by externalizing instance data to structured files before loading into the context.

\textbf{Long-Context Retrieval.}
\citet{liu2023lostmiddlelanguagemodels} show that LLMs struggle to retrieve from mid-context; RULER~\citep{hsieh2024rulerwhatsrealcontext} measures retrieval degradation using controlled tasks. Our experiments (\S\ref{sec:binding_bottleneck}) show that this retrieval degradation also explains binding failures in text-to-optimization, with multi-parameter retrieval exhibiting sharp accuracy cliffs as extraction failures compound.

\section{Text2Opt-Bench: Design and Evaluation}

\begin{figure}[t]
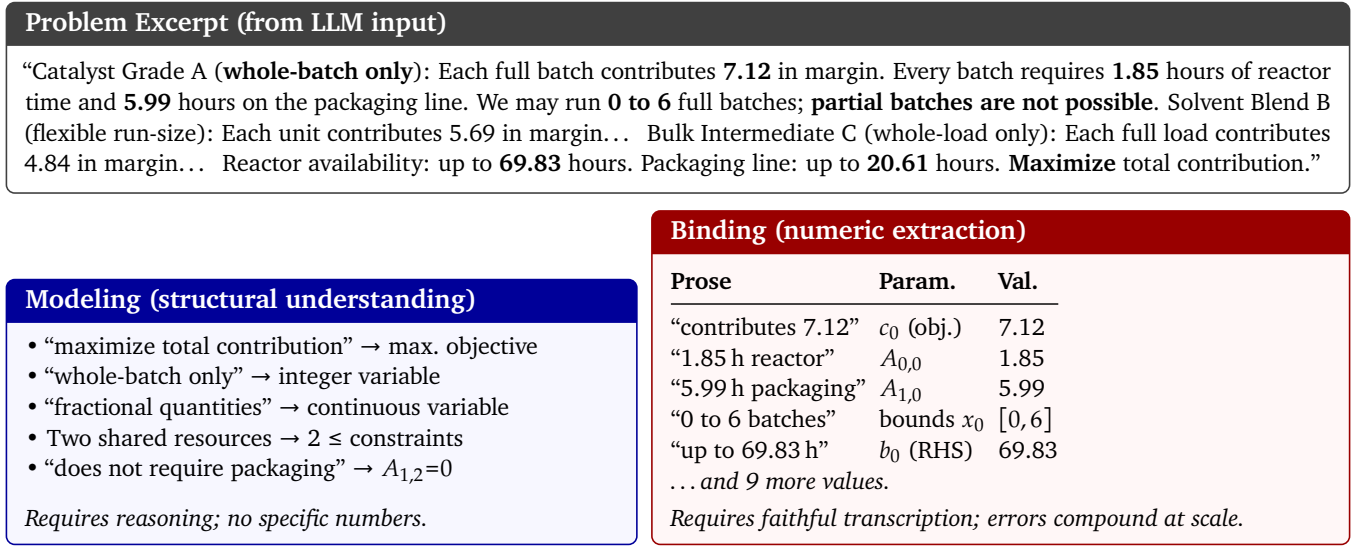

\centering
\begin{tcolorbox}[
    colback=white,
    colframe=black!75,
    title={\small\textbf{Problem Excerpt (from LLM input)}},
    fonttitle=\bfseries,
    boxrule=0.6pt,
    top=3pt, bottom=3pt, left=4pt, right=4pt
]
\footnotesize
``Catalyst Grade~A (\textbf{whole-batch only}): Each full batch contributes \textbf{7.12} in margin. Every batch requires \textbf{1.85} hours of reactor time and \textbf{5.99} hours on the packaging line. We may run \textbf{0 to 6} full batches; \textbf{partial batches are not possible}.
Solvent Blend~B (flexible run-size): Each unit contributes 5.69 in margin\ldots\;
Bulk Intermediate~C (whole-load only): Each full load contributes 4.84 in margin\ldots\;
Reactor availability: up to \textbf{69.83} hours. Packaging line: up to \textbf{20.61} hours. \textbf{Maximize} total contribution.''
\end{tcolorbox}

\vspace{-2pt}

\noindent
\begin{minipage}[t]{0.47\textwidth}
\begin{tcolorbox}[
    colback=blue!3,
    colframe=blue!60!black,
    title={\small\textbf{Modeling} (structural understanding)},
    fonttitle=\bfseries,
    boxrule=0.6pt,
    top=2pt, bottom=2pt, left=4pt, right=4pt
]
\footnotesize
\begin{itemize}[noitemsep,topsep=1pt,leftmargin=*,labelsep=3pt]
    \item ``maximize total contribution'' $\rightarrow$ max.\ objective
    \item ``whole-batch only'' $\rightarrow$ integer variable
    \item ``fractional quantities'' $\rightarrow$ continuous variable
    \item Two shared resources $\rightarrow$ 2\;$\leq$\;constraints
    \item ``does not require packaging'' $\rightarrow$ $A_{1,2}{=}0$
\end{itemize}
\smallskip
\emph{Requires reasoning; no specific numbers.}
\end{tcolorbox}
\end{minipage}%
\hfill
\begin{minipage}[t]{0.52\textwidth}
\begin{tcolorbox}[
    colback=red!3,
    colframe=red!60!black,
    title={\small\textbf{Binding} (numeric extraction)},
    fonttitle=\bfseries,
    boxrule=0.6pt,
    top=2pt, bottom=2pt, left=4pt, right=4pt
]
\footnotesize
\begin{tabular}{@{}l@{\;\;}l@{\;\;}l@{}}
\textbf{Prose} & \textbf{Param.} & \textbf{Val.} \\
\midrule
``contributes 7.12'' & $c_0$ (obj.) & 7.12 \\
``1.85\,h reactor'' & $A_{0,0}$ & 1.85 \\
``5.99\,h packaging'' & $A_{1,0}$ & 5.99 \\
``0 to 6 batches'' & bounds $x_0$ & $[0,6]$ \\
``up to 69.83\,h'' & $b_0$ (RHS) & 69.83 \\
\multicolumn{3}{@{}l@{}}{\textit{\ldots and 9 more values.}} \\
\end{tabular}
\smallskip

\emph{Requires faithful transcription; errors compound at scale.}
\end{tcolorbox}
\end{minipage}

\caption{Modeling vs.\ binding on a resource allocation instance.  \textbf{Modeling} selects the optimization structure (objective type, variable domains, constraints); \textbf{binding} extracts every numerical coefficient from prose. As instances scale, binding becomes the dominant failure mode.}
\label{fig:main_modeling_vs_binding}
\end{figure}

\label{sec:methodology}
Solving an optimization problem from natural language requires choosing the right mathematical structure and grounding that structure in the problem's numerical data. We formalize this decomposition first as it directly informs our benchmark design. Each problem category and evaluation mode is constructed to isolate one capability or the other.
\subsection{Problem Definition}

\label{subsec:decomposition}
We define \textit{text-to-optimization} as the task of producing executable solver code from a natural language description $D$. The description specifies both the problem's \textit{structure} (what to optimize, under what constraints) and its \textit{instance data} (the numerical coefficients, bounds, demands, and parameters). A correct solution requires two separable capabilities, as illustrated in Figure \ref{fig:main_modeling_vs_binding}.

\begin{itemize}[noitemsep,topsep=2pt,leftmargin=1.5em]
    \item \textbf{Modeling} $\mathcal{M}: (D^*, \theta) \to S$\,---\,given 
    a problem description $D^*$ and parameters $\theta$, select the 
    objective, constraints, and variable domains to produce executable 
    solver code $S$.
    \item \textbf{Binding} $\mathcal{B}: D \to \theta$\,---\, given a natural language description $D$, extract 
    concrete parameters $\theta$ (cost coefficients, capacity limits, 
    demand values, etc.)
\end{itemize}

An end-to-end approach performs both steps simultaneously: a 
single model maps $D$ directly to $S$, implicitly binding 
parameters while constructing the formulation (here $D^* = D$). 
A decomposed approach separates them: first extract 
$\theta = \mathcal{B}(D)$, then produce 
$S = \mathcal{M}(D^*, \theta)$, where $D^*$ may be $D$ itself 
or a structured representation.

Regardless of approach, these capabilities scale differently.
Modeling difficulty depends on the \textit{structural complexity} of the problem 
and is independent of instance scale.
The same structure must be selected regardless of the cardinality of $\theta$ (e.g., a transportation LP requires the same formulation whether it has 5 or 500 supply nodes). 
Binding difficulty grows with \textit{instance scale}, as each additional coefficient is an opportunity for transcription error. They are also empirically separable: varying instance scale at fixed structure isolates binding (\S\ref{subsec:binding_limit}); externalizing data isolates modeling (\S\ref{subsec:evaluation}). 
 
\vspace{-0.5em}
\subsection{Dataset Creation}
\label{subsec:pipeline}

\begin{figure}[t]
    \centering
    \includegraphics[width=0.95\textwidth]{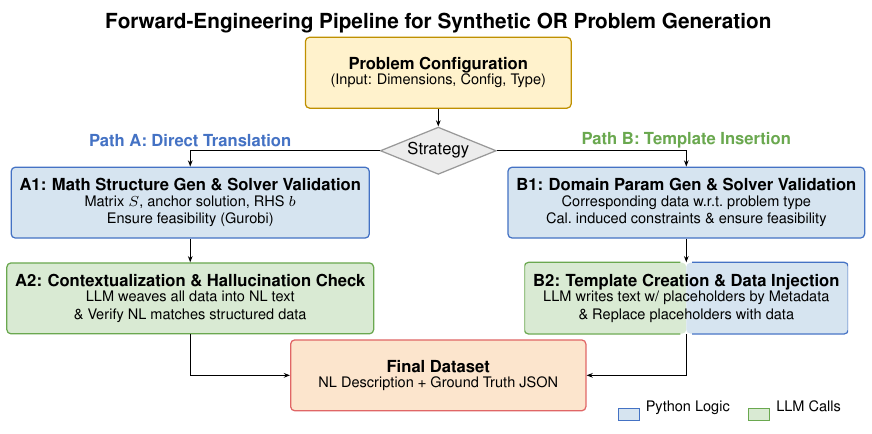}
    \caption{Text2Opt-Bench generation pipeline. Problems are constructed via forward engineering with solver verification, then described in natural language. Template-based insertion decouples linguistic complexity from data scale.}
    \label{fig:pipeline}
\end{figure}

Equipped with these definitions, we seek to build a dataset able to test models' abilities to handle modeling and binding. Rather than constructing constraints around a known solution (\emph{backward} engineering, as in OptMATH~\citep{lu2025optmathscalablebidirectionaldata}), we use a \emph{forward-engineering} framework (Figure~\ref{fig:pipeline}): (1)~simulate a \textit{world state}---business parameters, resource limits, and logical rules; (2)~derive the optimization structure and solve with an optimization solver\footnote{In this paper, we use Gurobi, a standard solver package; this choice is consistent with prior work~\citet{lu2025optmathscalablebidirectionaldata, berenbeim2025nlmoptimizer}.}; (3)~generate a natural language description grounded in the world state. This guarantees \textbf{feasibility by construction} and produces \textbf{semantically realistic} narratives. We adopt two complementary generation strategies: direct translation and template-based insertion. 

\textbf{Direct Translation.}
An LLM weaves all numerical coefficients directly into natural language prose. We use this for developing \textit{resource allocation} problems (LP/MILP, 2--20 variables), where the formulation requires minimal OR expertise but high faithfulness to the constraint values. Because the model must extract every coefficient from unstructured text, this category isolates binding difficulty from modeling difficulty (see Appendix~\ref{app:data_embedding} for an example). To confirm the binding difficulty of the constructed dataset, we analyze the failure modes of 9 models. Across all capable models, 60.4–92.3\% of resource allocation failures produce correct variable and constraint counts but wrong objective values; structural errors are near-zero (full results in Appendix~\ref{app:failure_modes}). 

\textbf{Template-Based Insertion.}
For structured problems requiring domain-specific modeling (for example, in scheduling, routing, and facility design), embedding all data in prose would exceed the model's effective binding capacity. Instead, we decouple language from data:
\begin{enumerate}[noitemsep,topsep=2pt]
    \item \textbf{Generate \& verify}: Create domain-specific parameters and solve with Gurobi.
    \item \textbf{Template}: LLMs generate natural language descriptions from data \emph{schema} only (dimensions, field names, no numeric values), with placeholders for data tables.
    \item \textbf{Insert}: Placeholders are filled deterministically with pipe-separated numerical data, enabling natural-language problem descriptions that scale to 1000+ variables.
\end{enumerate}

\textbf{Problem Categories.}
Text2Opt-Bench spans 12 categories organized into four tiers of increasing \emph{modeling} difficulty. Each template category includes 50 \textbf{small-tier} instances ($\leq$10K data tokens); three categories also include 50 \textbf{large-tier} instances ($\sim$30K tokens) with identical structure, isolating the effect of \emph{binding} scale. The pipeline is fully automated, so additional instances can be generated on demand.

The four tiers span increasing modeling difficulty: \textbf{Direct Translation} (Resource Allocation), \textbf{Template-Based} (Transportation, Disaster Response, JSSP, VRPTW, RCPSP), \textbf{Induced Constraint} (Facility Location, Power Transmission, Queuing/Staffing --- parameters derived from domain knowledge), and \textbf{Industrially-Motivated} (Stochastic Transportation, Multi-Objective Transportation, Modified Facility Location). Details are shown in Table~\ref{tab:main_results}.

\subsection{Evaluation Protocol}
\label{subsec:evaluation}

LLMs generate executable Gurobi Python code, which is run in a sandboxed subprocess. A response is correct iff the code (1) executes without error, (2) achieves optimal solver status, and (3) produces an objective value matching the ground truth within a relative tolerance of $10^{-4}$. All instances are feasible by construction --- otherwise, a wrong formulation could also return "infeasible" and be falsely marked correct.
Because coefficients are randomly generated continuous values, objective-value matching serves as an effective fingerprint: a wrong formulation is extremely unlikely to coincidentally produce the same optimum. We use objective matching rather than structural matching (e.g., variable or constraint counts) because many OR problems admit multiple valid formulations (details in Appendix~\ref{app:eval_validity}).

The benchmark naturally tests binding at three difficulty levels, from easiest to hardest.
\textbf{(1) Data-externalized} (BIND): numeric data lives in an external JSON file that the model can access. The model only needs to match constraint attributes to the corresponding keys and values in the file, making this the easiest binding setting.
\textbf{(2) Table-embedded} (template default): data appears as structured tables within the prompt. The model must locate and transcribe the correct entries from potentially large tables into code.
\textbf{(3) Prose-embedded} (direct translation): all coefficients are stated in natural-language sentences, requiring the model to parse numeric values from unstructured text. This is the hardest binding setting.
This design enables separate assessment of modeling vs.\ binding failures.

\textbf{BIND: a family of binding-aware techniques}.
BIND is a family of techniques that relieve the model of coefficient transcription. Its inference-time component, \emph{data offloading}, externalizes all numeric data (e.g. cost matrices) to a JSON file. The model receives: (1) the structural problem description (objectives and constraints in natural language), (2) the data schema with dimensions and types, and (3) a file path. This forces the model to bind programmatically via \texttt{json.load()} rather than transcribing coefficients from the prompt. Data offloading runs either as a single pass or with oracle-guided iterative refinement (here, using $\le$3 rounds). We treat the refinement variant as the main inference result and the single pass variant as an ablation that serves to isolate pure transcription removal (\S\ref{subsubsec:inference}) as diagnostic for recoverable accuracy. 

\section{The Binding Bottleneck}
\label{subsec:binding_limit}

We observe that binding is the dominant bottleneck on most categories (a per-category split is in Table~\ref{tab:category_regimes}). We first show that accuracy collapses as the data scale increases, even when the formulation structure is fixed (Table~\ref{tab:binding_degrade}), then benchmark 9 models across the dataset (\S\ref{subsec:main_results}), and finally confirm via RULER retrieval tasks that this reflects a general limitation in context-processing (\S\ref{sec:binding_bottleneck}).
\begin{figure}[t]
    \centering
    \includegraphics[width=\textwidth]{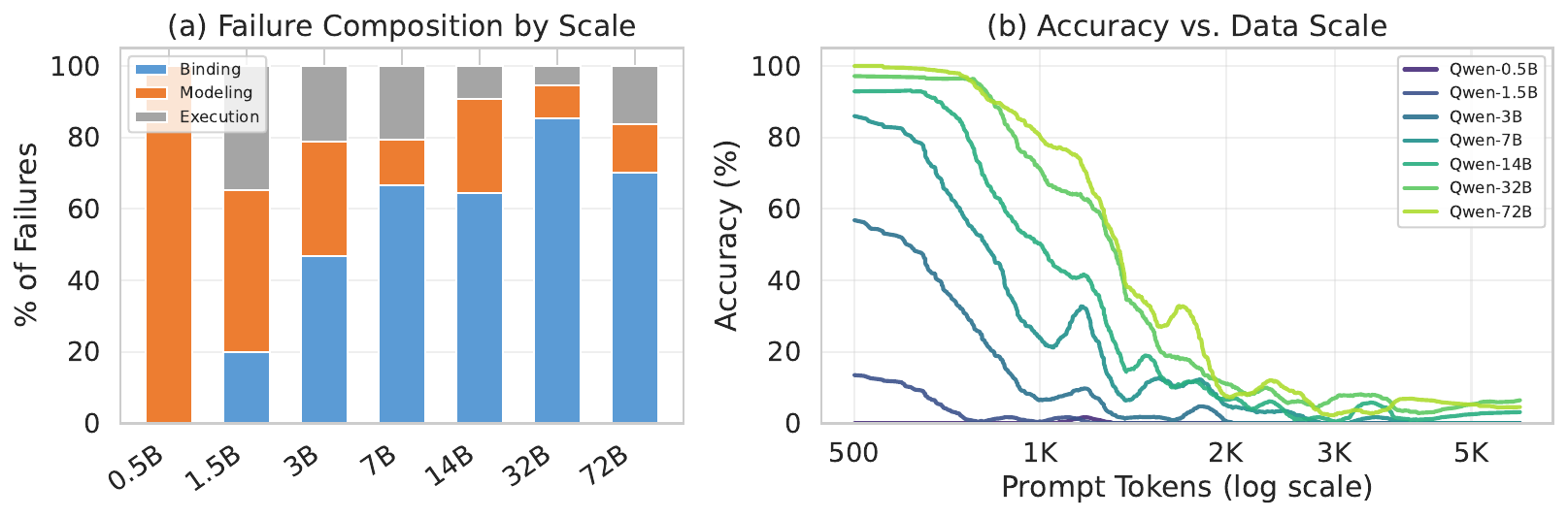}
    \caption{\textbf{(a)} Failure composition by model scale on resource allocation (1,012 problems). As model size grows, binding errors increasingly make up a significant proportion of failures. \textbf{(b)} Each model exhibits an \emph{effective binding limit} beyond which accuracy sharply declines. Curves are smoothed with a Gaussian-weighted rolling average.}
    \label{fig:context_scaling}
\end{figure}

\textbf{Direct translation.}
Figure~\ref{fig:context_scaling} presents the relationship between model scale, data scale, and binding failures on all 1,012 resource allocation problems, which isolates binding as discussed in \S\ref{subsec:pipeline}, across the Qwen2.5 family (0.5B--72B) to control for architectural differences. Panel (b) shows accuracy as a function of prompt token length.

We observe three primary trends: \textbf{(1)~Binding failures dominate at scale:} Panel~(a) shows a phase transition in failure composition: at 0.5B, nearly all failures are modeling errors (the model cannot formulate LPs); by 32B, 86\% of failures are binding errors---correct formulation structure but wrong coefficients. \textbf{(2)~Accuracy declines sharply with instance scale:}  Panel~(b) shows that accuracy drops as the size of the optimization problem grows. This confirms that the advertised context window (128k for Qwen-2.5 family) is far larger than the \textit{effective} window for dense numerical tasks, aligning with recent findings on context scaling limits~\citep{shi2026intrinsicentropycontextlength, zhou2025gsminfinitellmsbehaveinfinitely, liu2023lostmiddlelanguagemodels}. \textbf{(3)~Model-specific thresholds:} Larger models maintain accuracy on longer prompts. This shows a clear correlation between parameter count and effective context length.

\begin{table}[t]
\centering
\caption{Binding degradation: small (n=50, ${\leq}$10K tokens) vs.\ large (n=50, 23--35K tokens) on three binding-limited categories. Same formulation structure, only data scale changes.}
\label{tab:binding_degrade}
\small
\begin{tabular}{l|rr|cc|cc}
\toprule
 & \multicolumn{2}{c|}{\textbf{Avg Tokens}} & \multicolumn{2}{c|}{\textbf{GPT-5}} & \multicolumn{2}{c}{\textbf{GPT-5-Nano}} \\
\textbf{Category} & \textbf{Small} & \textbf{Large} & \textbf{S} & \textbf{L} & \textbf{S} & \textbf{L} \\
\midrule
Transportation & 1.4K & 23K & 100 & 90 \tr{$-$10} & 100 & 32 \tr{$-$68} \\
Multi-Obj T. & 3.6K & 35K & 70 & 48 \tr{$-$22} & 60 & 0 \tr{$-$60} \\
Queue/Staff. & 5.4K & 34K & 80 & 66 \tr{$-$14} & 56 & 0 \tr{$-$56} \\
\midrule
\textbf{Average} & & & 83 & 68 \tr{$-$15} & 72 & 11 \tr{$-$61} \\
\bottomrule
\end{tabular}
\end{table}

\textbf{Template binding degradation.}
Table~\ref{tab:binding_degrade} confirms this on a set of structured problems. Accuracy degrades from 83\% to 68\% (GPT-5) and from 72\% to 11\% (GPT-5-Nano) when scaling from ${\leq}$10K to $\sim$30K data tokens at identical structure. Transportation is the clearest case: both models achieve 100\% on small-tier instances, ruling out any formulation difficulty; the drop in GPT-5-Nano's accuracy to 32\% on large instances is therefore attributable purely to binding scale, consistent with the multi-key retrieval cliff observed in RULER (\S\ref{sec:binding_bottleneck}).

\subsection{Model and Scale Comparison}
\label{subsec:main_results}

We evaluate Text2Opt-Bench across 9 models on the main benchmark (Table~\ref{tab:main_results}), with additional scale analysis across the Qwen-2.5 family (0.5B–72B). All problems' descriptions are generated using GPT-5, at a cost of ${\sim}\$0.03$ (template) to ${\sim}\$0.10$ (direct translation) per instance. We also evaluated GPT-5; data leakage concerns are discussed in Appendix~\ref{app:description_robustness}.

Table~\ref{tab:main_results} presents pass@1 accuracy on the 550 small-tier template problems (50 per category). We measure correctness as described in \S\ref{subsec:evaluation}. This table reveals several patterns. \textbf{(1) Frontier models are closely matched}: Claude Sonnet~4.6, Opus~4.6, and GPT-5 achieve 84--90\% on both resource allocation and template problems. 
\textbf{(2) Reasoning models do not outperform standard models}: DeepSeek-R1 performs similarly to DeepSeek-V3.2, suggesting that chain-of-thought reasoning does not address the binding bottleneck (Appendix~\ref{app:prompting} confirms that prompting strategies also fail to help). 
\textbf{(3) Small models lack modeling skills}: Qwen2.5-7B achieves 0\% across many categories.

\begin{table}[t]
\caption{Text2Opt-Bench pass@1 accuracy (\%). Template categories: 50 small-tier instances each. Resource allocation: 248 eval-subset instances. Best per row in \textbf{bold}. $\dagger$: 50 additional large-tier instances ($\sim$30K data tokens) for binding stress tests (\S\ref{subsec:binding_limit}).}
\label{tab:main_results}
\begin{center}
\scriptsize
\setlength{\tabcolsep}{2.5pt}
\begin{tabular}{llcc|ccc|cc|c|c|cc}
\toprule
 &  & & & \multicolumn{3}{c|}{\textit{Frontier}} & \multicolumn{2}{c|}{\textit{Reasoning}} & \textit{Open-Large} & \textit{Close-Mid} & \multicolumn{2}{c}{\textit{Open-Other}} \\
\textbf{Category} & \textbf{Problem Type} & \textbf{Form.} & \shortstack{\textbf{Variable} \\ \textbf{Count}} & \rotatebox{70}{Sonnet 4.6} & \rotatebox{70}{Opus 4.6} & \rotatebox{70}{GPT-5} & \rotatebox{70}{o4-mini} & \rotatebox{70}{DS-R1} & \rotatebox{70}{DS-V3.2} & \rotatebox{70}{GPT-5-Nano} & \rotatebox{70}{Llama3.3-70B} & \rotatebox{70}{Qwen2.5-7B} \\
\midrule
Direct Tran. & Resource Alloc. & LP/MILP & 2--20 & 84.7 & \textbf{89.9} & 87.9 & 80.2 & 80.6 & 79.0 & 49.2 & 49.6 & 13.3 \\
\midrule
\multirow{5}{*}{\shortstack{Template \\ Based}} & Transportation$^\dagger$ & LP & 9--625 & \textbf{100} & 98 & \textbf{100} & \textbf{100} & \textbf{100} & \textbf{100} & \textbf{100} & 88 & 38 \\
 & Disaster Resp. & MILP & 30--792 & \textbf{96} & \textbf{96} & 86 & 94 & 78 & 90 & 62 & 30 & 0 \\
 & JSSP & MILP & 19--365 & 98 & \textbf{100} & 90 & 96 & 96 & 96 & 82 & 0 & 0 \\
 & VRPTW & MILP & 41--419 & 50 & 38 & \textbf{70} & 34 & 34 & 22 & 2 & 0 & 0 \\
 & RCPSP & MILP & 26--181 & \textbf{100} & 96 & 88 & 82 & 34 & 62 & 26 & 0 & 0 \\
\midrule
\multirow{3}{*}{\shortstack{Induced \\ Constraint}} & Facility Loc. & MILP & 18--980 & 98 & \textbf{100} & \textbf{100} & 98 & 94 & 98 & 90 & 98 & 6 \\
 & Power Trans. & MIQP & 18--360 & 64 & 88 & \textbf{98} & 70 & 64 & 54 & 50 & 16 & 0 \\
 & Queuing/Staff.$^\dagger$ & NLP & 36--2.6K & \textbf{98} & 92 & 80 & 76 & 70 & 66 & 56 & 10 & 0 \\
\midrule
\multirow{3}{*}{\shortstack{Industrially \\ Motivated}} & Stoch.\ Transp. & MILP & 172--1.4K & 62 & 62 & \textbf{70} & 66 & 60 & 18 & 32 & 6 & 0 \\
 & Multi-Obj T.$^\dagger$ & MILP & 40--896 & \textbf{98} & 88 & 70 & 68 & 76 & 86 & 60 & 42 & 4 \\
 & Mod.\ Fac.\ Loc. & MILP & 28--390 & \textbf{100} & 96 & 96 & \textbf{100} & 96 & \textbf{100} & 90 & 96 & 2 \\
\midrule
\multicolumn{4}{l|}{\textbf{Template Avg.}} & \textbf{87.6} & 86.7 & 86.2 & 80.4 & 72.9 & 72.0 & 59.1 & 35.1 & 4.5 \\
\bottomrule
\end{tabular}
\end{center}
\end{table}

\subsection{Retrieval Failures Beyond Optimization}
\label{sec:binding_bottleneck}

To isolate the retrieval component of binding from OR-specific knowledge, we evaluate the Qwen2.5 family (0.5B--32B) on four tasks adapted from the RULER long-context benchmark~\citep{hsieh2024rulerwhatsrealcontext}: single-key retrieval (analogous to reading \texttt{demand[j]}), multi-key retrieval (binding all coefficients in a constraint), multi-value retrieval (reading a data column), and aggregation (assembling an objective from scattered data). We harden RULER with distractor keys and scale difficulty by context length (1K--32K tokens). All tasks use strict exact-match: every requested value must be correct. Full details are in Appendix~\ref{app:ruler}.

\begin{figure}[t]
    \centering
    \includegraphics[width=\textwidth]{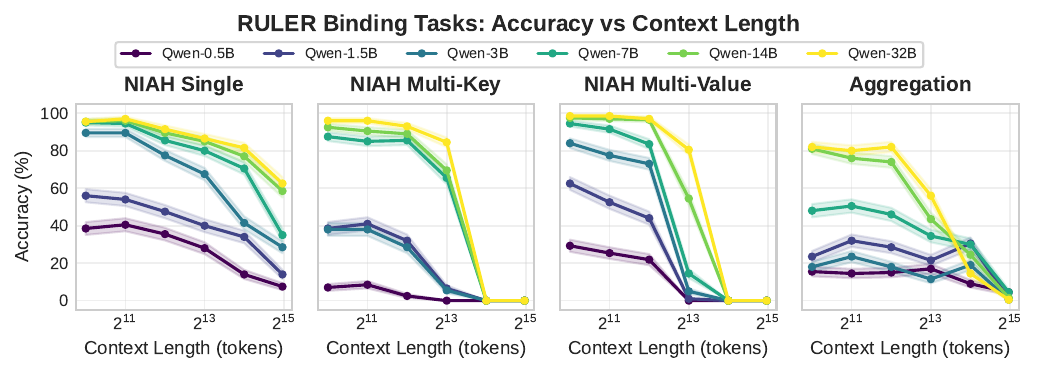}
    \caption{Accuracy on four RULER binding tasks across Qwen-2.5 sizes (0.5B--32B). Strict exact-match scoring; 200 samples per task per context length. Multi-binding tasks exhibit sharp cliffs as individual retrieval failures compound multiplicatively.}
    \label{fig:ruler_binding}
\end{figure}

Figure~\ref{fig:ruler_binding} reveals two findings. First, \emph{every} model degrades as context grows---even Qwen-32B's average score drops from 90\% (1K) to 16\% (32K). Second, degradation depends on the number of simultaneous bindings: single-key retrieval degrades gradually (32B: $96\% \to 63\%$), whereas multi-key and multi-value retrieval collapse from $>$90\% to 0\% between 8K and 16K tokens. This cliff is consistent with per-binding failure rates that compound multiplicatively ($p^k$), confirming that the binding bottleneck reflects a general retrieval limitation, instead of an OR-specific deficit.

\textbf{Summary.} The evidence above cleanly separates two failure regimes. For \textbf{binding-limited} categories (transportation, facility location, JSSP, queuing/staffing), BIND recovers most failures: GPT-5 reaches 98--100\%, confirming that residual errors were transcription failures. For \textbf{modeling-limited} categories (VRPTW, stochastic transportation, power transmission), BIND provides smaller or no gains---these failures reflect structural errors such as incorrect subtour elimination or mis-formulated chance constraints (see Appendix~\ref{app:power_bind_regression} for details).

\section{Mitigating Binding Failures}
\label{subsec:mitigating}

Having established binding as the primary bottleneck, we now want to investigate how this can be addressed. We consider two complementary approaches: inference-time strategies (\S\ref{subsubsec:inference}), and training-time strategies that specialize a model for binding (\S\ref{subsubsec:training}).

\subsection{Inference: BIND and Test-Time Compute}
\label{subsubsec:inference}
We evaluate test-time compute (TTC) strategies that trade additional inference cost for higher accuracy, including repeated sampling~\citep{brown2024largelanguagemonkeysscaling, snell2024scalingllmtesttimecompute}, iterative repair~\citep{chen2023teachinglargelanguagemodels}, and our binding-aware data offloading (BIND).

Figure~\ref{fig:bind_pareto} and Table~\ref{tab:ttc_full} compare these strategies across seven models from three families on 550 template problems (Llama3.3-70B and Qwen2.5-7B are excluded due to insufficient modeling ability; their results are in Appendix~\ref{app:ttc_full}). BIND runs data offloading with $\le$3 rounds of oracle-guided refinement (this is the same oracle signal as the inline \textbf{Repair@5} baseline). We add \textbf{pass@5} as a parallel-sampling baseline. A single-pass ablation (no refinement) is used to isolate pure transcription removal (App.~\ref{app:bind_ablation_padding}).

BIND reaches near-ceiling accuracy at low token cost (GPT-5 95.8\%, Claude Opus 98.7\%), matching Repair@5 and pass@5 while spending far fewer tokens as offloading shrinks the per-round prompt (Figure~\ref{fig:bind_pareto}). The single-pass ablation alone recovers +12 to +27pp on binding-limited categories, confirming that the binding bottleneck is the primary recoverable failure mode. Refinement closes the remaining gap in modeling-limited categories. BIND also outperforms a search-over-formulations baseline (SolverLLM/MCTS-oracle@5) at 5--12$\times$ lower token cost (App.~\ref{app:mcts_baseline}).

BIND costs ${\sim}3$--$5$K tokens (summed over its refinement rounds), comparable to pass@1 (${\sim}4$K) and well below inline repair (Repair@5 6--12K) and pass@5 (${\sim}20$K). We show the full per-model breakdown including token costs in Table~\ref{tab:ttc_full} (Appendix~\ref{app:ttc_full}).

\textbf{BIND per-category analysis.}
BIND consistently improves all capable models: GPT-5 gains +9.6pp, Sonnet 4.6 +8.6pp, DeepSeek-R1 +11.6pp, and GPT-5-Nano +23.3pp. The largest gains are on data-heavy categories (+56pp for GPT-5-Nano on stochastic transportation).  However, as expected, BIND does not compensate for missing \emph{modeling} ability: Qwen2.5-7B's accuracy remains 0\% on all structurally complex categories (VRPTW, RCPSP, stochastic). Per-category results are found in Appendix~\ref{app:ttc_full} Table~\ref{tab:bind}. Appendix~\ref{app:binding_case} provides a detailed case study of binding failures. An additional benchmark-robustness check shows model accuracy is insensitive to which model authored the prose (App.~\ref{app:description_robustness}).

\subsection{Training binding-specific models is effective}
\label{subsubsec:training}
If binding is the bottleneck, then a model trained \emph{only} to bind should outperform one trained end-to-end. We test this with a two-phase pipeline: a fine-tuned binding model produces structured JSON, and a separate solver stage—an untrained LLM or deterministic template code—constructs the Gurobi program. We compare against standard supervised finetuning and also reinforcement learning via GRPO \citep{shao2024deepseekmathpushinglimitsmathematical}.

Table~\ref{tab:binding-pipeline} reports results. Across all three categories, the 7B binding specialist outperforms end-to-end SFT: 58.1\% vs.\ 51.2\% (resource allocation), 100\% vs.\ 96\% (JSSP), and 96.0\% vs.\ 88.0\% (transportation). GRPO underperforms SFT, and adding denser reward signals (hierarchical partial credit) further degrades accuracy as the model exploits intermediate reward gates (see Appendix~\ref{app:grpo} and~\ref{app:rl_ablations} for a full study). In-distribution accuracy is near-perfect (96--100\%) across categories, and a 1.5B binding specialist already matches 7B end-to-end SFT on resource allocation. Fixed-schema categories (transportation, JSSP) generalize well OOD (91.7--100\%), while free-form categories like resource allocation require training coverage closer to the target distribution (Appendix~\ref{app:ood-shift}). A single multi-task binding specialist trained jointly on all three categories is competitive with the per-category specialists (App.~\ref{app:multitask_specialist}). These results reinforce the modeling--binding decomposition: isolating the bottleneck task yields both stronger performance and better parameter efficiency than joint training, with SFT's dense token-level supervision proving more suited to faithful transcription than RL's sparse outcome-based reward.

\begin{table}[t]
\caption{Accuracy of two-phase binding pipeline vs.\ end-to-end training across three categories.
Phase~1 binding specialists are Qwen2.5-7B-Instruct (full SFT)\@.
Phase~2 uses an untrained Qwen2.5-7B for resource allocation and deterministic
template code for transportation and JSSP\@.}
\label{tab:binding-pipeline}
\centering
\small
\begin{tabular}{ll ccc}
\toprule
Category & System & All & In-distribution & OOD \\
\midrule
\multirow{5}{*}{\shortstack[l]{Resource Alloc.\\[-1pt]{\scriptsize Phase 2: Qwen-7B}}} 
  & Ground-truth $\to$ Qwen-7B & 100 & 100 & 100 \\
  & 7B binding spec.\ $\to$ Qwen-7B & \textbf{58.1} & 99.2 & 11.2 \\
  & 1.5B binding spec.\ $\to$ Qwen-7B & 51.2 & 94.7 & 1.7 \\
  & 7B SFT & 51.2 & 88.6 & 8.6 \\
  & 7B GRPO \footnotemark & 44.0 & 76.5 & 6.9 \\ 
\midrule
\multirow{3}{*}{\shortstack[l]{Transportation\\[-1pt]{\scriptsize Phase 2: Template}}} 
  & Ground-truth $\to$ Template & 100 & 100 & 100 \\
  & 7B binding spec.\ $\to$ Template & \textbf{96.0} & 100 & \textbf{91.7} \\
  & 7B SFT & 88.0 & 100 & 75.0 \\
\midrule
\multirow{3}{*}{\shortstack[l]{JSSP\\[-1pt]{\scriptsize Phase 2: Template}}} 
  & Ground-truth $\to$ Template & 100 & 100 & 100 \\
  & 7B binding spec.\ $\to$ Template & \textbf{100} & 100 & \textbf{100} \\
  & 7B SFT & 96.0 & 100 & 92.0 \\
\bottomrule
\end{tabular}
\end{table}
\footnotetext{RL requires a reward signal from executing generated code;
at 7B scale, the base model has 0\% base accuracy for JSSP. Thus, we opted not to include RL baselines for the categories in the table.}

\section{Conclusion}
We presented Text2Opt-Bench, a benchmark of 12 solver-verified optimization categories, and showed that instance binding is the dominant bottleneck on most categories for frontier LLMs. BIND, training, and controlled retrieval experiments on RULER tasks all converge on this conclusion, while modeling limitations still exist for structurally complex problems (VRPTW, power transmission). SFT outperforms RL at 7B scale; and binding specialists outperform end-to-end SFT across three categories—all consistent with binding as the bottleneck.

\textbf{Limitations:}
Our benchmark covers mathematical programming formulations solvable by Gurobi but does not cover combinatorial optimization requiring heuristic or metaheuristic approaches. The fine-tuning study covers three categories (resource allocation, transportation, JSSP) at 7B scale; extension to structurally complex categories (VRPTW, stochastic transportation), larger models, and to messy descriptions, tables, or PDFs in realistic settings, remains future work.

\section*{Acknowledgments}
We are grateful for the support of the National Science Foundation (NSF) (CCF2106707), the Defense
Advanced Research Projects Agency (DARPA Young Faculty Award), the Wisconsin Alumni Research Foundation (WARF).

\section*{Ethics Statement}
This work uses GPT-5 to generate natural language problem descriptions for benchmark instances (Section~\ref{subsec:pipeline}). The generated solution is solver-verified via Gurobi; no LLM is used for evaluation or scoring. No human subjects, personally identifiable information, or sensitive data are involved in this work.

\bibliography{references}

\begin{thebibliography}{33}
\providecommand{\natexlab}[1]{#1}
\providecommand{\url}[1]{\texttt{#1}}
\expandafter\ifx\csname urlstyle\endcsname\relax
  \providecommand{\doi}[1]{doi: #1}\else
  \providecommand{\doi}{doi: \begingroup \urlstyle{rm}\Url}\fi

\bibitem[AhmadiTeshnizi et~al.(2024)AhmadiTeshnizi, Gao, and Udell]{ahmaditeshnizi2024optimusscalableoptimizationmodeling}
Ali AhmadiTeshnizi, Wenzhi Gao, and Madeleine Udell.
\newblock Optimus: Scalable optimization modeling with (mi)lp solvers and large language models, 2024.
\newblock URL \url{https://arxiv.org/abs/2402.10172}.

\bibitem[Berenbeim et~al.(2025)Berenbeim, McNeil, Williams, and Bastian]{berenbeim2025nlmoptimizer}
Alexander~Michael Berenbeim, Ryan McNeil, Timeo Williams, and Nathaniel~D. Bastian.
\newblock {NLMO}ptimizer: A neurosymbolic framework and benchmark for operations research optimization problems from natural language, 2025.
\newblock URL \url{https://openreview.net/forum?id=skctEx59f2}.

\bibitem[Brown et~al.(2024)Brown, Juravsky, Ehrlich, Clark, Le, Ré, and Mirhoseini]{brown2024largelanguagemonkeysscaling}
Bradley Brown, Jordan Juravsky, Ryan Ehrlich, Ronald Clark, Quoc~V. Le, Christopher Ré, and Azalia Mirhoseini.
\newblock Large language monkeys: Scaling inference compute with repeated sampling, 2024.
\newblock URL \url{https://arxiv.org/abs/2407.21787}.

\bibitem[Chen et~al.(2023{\natexlab{a}})Chen, Ma, Wang, and Cohen]{chen2023programthoughtspromptingdisentangling}
Wenhu Chen, Xueguang Ma, Xinyi Wang, and William~W. Cohen.
\newblock Program of thoughts prompting: Disentangling computation from reasoning for numerical reasoning tasks, 2023{\natexlab{a}}.
\newblock URL \url{https://arxiv.org/abs/2211.12588}.

\bibitem[Chen et~al.(2023{\natexlab{b}})Chen, Lin, Schärli, and Zhou]{chen2023teachinglargelanguagemodels}
Xinyun Chen, Maxwell Lin, Nathanael Schärli, and Denny Zhou.
\newblock Teaching large language models to self-debug, 2023{\natexlab{b}}.
\newblock URL \url{https://arxiv.org/abs/2304.05128}.

\bibitem[Chen et~al.(2026)Chen, Cheng, Sun, Ling, and Ge]{chen2026optenginebenchmarkinglimitsllms}
Yitian Chen, Cheng Cheng, Yinan Sun, Zi~Ling, and Dongdong Ge.
\newblock Opt-engine: Benchmarking the limits of llms in optimization modeling via complexity scaling, 2026.
\newblock URL \url{https://arxiv.org/abs/2601.19924}.

\bibitem[Gao et~al.(2023)Gao, Madaan, Zhou, Alon, Liu, Yang, Callan, and Neubig]{gao2023palprogramaidedlanguagemodels}
Luyu Gao, Aman Madaan, Shuyan Zhou, Uri Alon, Pengfei Liu, Yiming Yang, Jamie Callan, and Graham Neubig.
\newblock Pal: Program-aided language models, 2023.
\newblock URL \url{https://arxiv.org/abs/2211.10435}.

\bibitem[Goldie et~al.(2025)Goldie, Mirhoseini, Zhou, Cai, and Manning]{goldie2025syntheticdatageneration}
Anna Goldie, Azalia Mirhoseini, Hao Zhou, Irene Cai, and Christopher~D. Manning.
\newblock Synthetic data generation and multi-step rl for reasoning and tool use, 2025.
\newblock URL \url{https://arxiv.org/abs/2504.04736}.

\bibitem[Hsieh et~al.(2024)Hsieh, Sun, Kriman, Acharya, Rekesh, Jia, Zhang, and Ginsburg]{hsieh2024rulerwhatsrealcontext}
Cheng-Ping Hsieh, Simeng Sun, Samuel Kriman, Shantanu Acharya, Dima Rekesh, Fei Jia, Yang Zhang, and Boris Ginsburg.
\newblock Ruler: What's the real context size of your long-context language models?, 2024.
\newblock URL \url{https://arxiv.org/abs/2404.06654}.

\bibitem[Huang et~al.(2025{\natexlab{a}})Huang, Tang, Hu, Jiang, Zheng, Ge, Wang, and Wang]{Huang_2025}
Chenyu Huang, Zhengyang Tang, Shixi Hu, Ruoqing Jiang, Xin Zheng, Dongdong Ge, Benyou Wang, and Zizhuo Wang.
\newblock Orlm: A customizable framework in training large models for automated optimization modeling.
\newblock \emph{Operations Research}, 73\penalty0 (6):\penalty0 2986–3009, November 2025{\natexlab{a}}.
\newblock ISSN 1526-5463.
\newblock \doi{10.1287/opre.2024.1233}.
\newblock URL \url{http://dx.doi.org/10.1287/opre.2024.1233}.

\bibitem[Huang et~al.(2025{\natexlab{b}})Huang, Shen, Hu, Gao, and Wang]{huang2025llmsmathematicalmodelingbridging}
Xuhan Huang, Qingning Shen, Yan Hu, Anningzhe Gao, and Benyou Wang.
\newblock Llms for mathematical modeling: Towards bridging the gap between natural and mathematical languages, 2025{\natexlab{b}}.
\newblock URL \url{https://arxiv.org/abs/2405.13144}.

\bibitem[Jiang et~al.(2025)Jiang, Shu, Qian, Lu, Zhou, Zhou, and Yu]{jiang2025llmoptlearningdefinesolve}
Caigao Jiang, Xiang Shu, Hong Qian, Xingyu Lu, Jun Zhou, Aimin Zhou, and Yang Yu.
\newblock Llmopt: Learning to define and solve general optimization problems from scratch, 2025.
\newblock URL \url{https://arxiv.org/abs/2410.13213}.

\bibitem[Kwon et~al.(2023)Kwon, Li, Zhuang, Sheng, Zheng, Yu, Gonzalez, Zhang, and Stoica]{kwon2023efficientmemorymanagementlarge}
Woosuk Kwon, Zhuohan Li, Siyuan Zhuang, Ying Sheng, Lianmin Zheng, Cody~Hao Yu, Joseph~E. Gonzalez, Hao Zhang, and Ion Stoica.
\newblock Efficient memory management for large language model serving with pagedattention, 2023.
\newblock URL \url{https://arxiv.org/abs/2309.06180}.

\bibitem[Li et~al.(2025)Li, Zhao, Yu, Liu, Cheng, Chen, Chen, Chen, Zhao, and Chen]{li2025solverllmleveragingtesttimescaling}
Dong Li, Xujiang Zhao, Linlin Yu, Yanchi Liu, Wei Cheng, Zhengzhang Chen, Zhong Chen, Feng Chen, Chen Zhao, and Haifeng Chen.
\newblock Solverllm: Leveraging test-time scaling for optimization problem via llm-guided search, 2025.
\newblock URL \url{https://arxiv.org/abs/2510.16916}.

\bibitem[Liu et~al.(2025)Liu, Fan, Jiang, Ding, Hu, Zhang, Shi, Weng, Chen, Chen, Huang, Zhang, Zhao, Yan, and He]{liu2025synlogicsynthesizingverifiablereasoning}
Junteng Liu, Yuanxiang Fan, Zhuo Jiang, Han Ding, Yongyi Hu, Chi Zhang, Yiqi Shi, Shitong Weng, Aili Chen, Shiqi Chen, Yunan Huang, Mozhi Zhang, Pengyu Zhao, Junjie Yan, and Junxian He.
\newblock Synlogic: Synthesizing verifiable reasoning data at scale for learning logical reasoning and beyond, 2025.
\newblock URL \url{https://arxiv.org/abs/2505.19641}.

\bibitem[Liu et~al.(2023)Liu, Lin, Hewitt, Paranjape, Bevilacqua, Petroni, and Liang]{liu2023lostmiddlelanguagemodels}
Nelson~F. Liu, Kevin Lin, John Hewitt, Ashwin Paranjape, Michele Bevilacqua, Fabio Petroni, and Percy Liang.
\newblock Lost in the middle: How language models use long contexts, 2023.
\newblock URL \url{https://arxiv.org/abs/2307.03172}.

\bibitem[Lu et~al.(2025)Lu, Xie, Wu, Ren, Chen, and Wen]{lu2025optmathscalablebidirectionaldata}
Hongliang Lu, Zhonglin Xie, Yaoyu Wu, Can Ren, Yuxuan Chen, and Zaiwen Wen.
\newblock Optmath: A scalable bidirectional data synthesis framework for optimization modeling, 2025.
\newblock URL \url{https://arxiv.org/abs/2502.11102}.

\bibitem[Mostajabdaveh et~al.(2025)Mostajabdaveh, Yu, Dash, Ramamonjison, Byusa, Carenini, Zhou, and Zhang]{mostajabdaveh2025evaluatingllmreasoningoperations}
Mahdi Mostajabdaveh, Timothy~T. Yu, Samarendra Chandan~Bindu Dash, Rindranirina Ramamonjison, Jabo~Serge Byusa, Giuseppe Carenini, Zirui Zhou, and Yong Zhang.
\newblock Evaluating llm reasoning in the operations research domain with orqa, 2025.
\newblock URL \url{https://arxiv.org/abs/2412.17874}.

\bibitem[Ramamonjison et~al.(2022)Ramamonjison, Yu, Li, Li, Carenini, Ghaddar, He, Mostajabdaveh, Banitalebi-Dehkordi, Zhou, and Zhang]{pmlr-v220-ramamonjison23a}
Rindranirina Ramamonjison, Timothy Yu, Raymond Li, Haley Li, Giuseppe Carenini, Bissan Ghaddar, Shiqi He, Mahdi Mostajabdaveh, Amin Banitalebi-Dehkordi, Zirui Zhou, and Yong Zhang.
\newblock Nl4opt competition: Formulating optimization problems based on their natural language descriptions.
\newblock In Marco Ciccone, Gustavo Stolovitzky, and Jacob Albrecht, editors, \emph{Proceedings of the NeurIPS 2022 Competitions Track}, volume 220 of \emph{Proceedings of Machine Learning Research}, pages 189--203. PMLR, 28 Nov--09 Dec 2022.
\newblock URL \url{https://proceedings.mlr.press/v220/ramamonjison23a.html}.

\bibitem[Seegmiller et~al.(2025)Seegmiller, Mehta, Saha, Tao, Oraby, Gupta, Chung, Bansal, and Peng]{seegmiller2025flamesimprovingllmmath}
Parker Seegmiller, Kartik Mehta, Soumya Saha, Chenyang Tao, Shereen Oraby, Arpit Gupta, Tagyoung Chung, Mohit Bansal, and Nanyun Peng.
\newblock Flames: Improving llm math reasoning via a fine-grained analysis of the data synthesis pipeline, 2025.
\newblock URL \url{https://arxiv.org/abs/2508.16514}.

\bibitem[Shao et~al.(2024)Shao, Wang, Zhu, Xu, Song, Bi, Zhang, Zhang, Li, Wu, and Guo]{shao2024deepseekmathpushinglimitsmathematical}
Zhihong Shao, Peiyi Wang, Qihao Zhu, Runxin Xu, Junxiao Song, Xiao Bi, Haowei Zhang, Mingchuan Zhang, Y.~K. Li, Y.~Wu, and Daya Guo.
\newblock Deepseekmath: Pushing the limits of mathematical reasoning in open language models, 2024.
\newblock URL \url{https://arxiv.org/abs/2402.03300}.

\bibitem[Shen et~al.(2026)Shen, Guo, Wan, Yang, Zhang, Huang, Song, Zhang, and Sun]{shen2026proopfbenchmarkingimprovingllms}
Chao Shen, Zihan Guo, Xu~Wan, Zhenghao Yang, Yifan Zhang, Wengi Huang, Jie Song, Zongyan Zhang, and Mingyang Sun.
\newblock Proopf: Benchmarking and improving llms for professional-grade power systems optimization modeling, 2026.
\newblock URL \url{https://arxiv.org/abs/2602.03070}.

\bibitem[Sheng et~al.(2025)Sheng, Zhang, Ye, Wu, Zhang, Zhang, Peng, Lin, and Wu]{Sheng_2025}
Guangming Sheng, Chi Zhang, Zilingfeng Ye, Xibin Wu, Wang Zhang, Ru~Zhang, Yanghua Peng, Haibin Lin, and Chuan Wu.
\newblock Hybridflow: A flexible and efficient rlhf framework.
\newblock In \emph{Proceedings of the Twentieth European Conference on Computer Systems}, EuroSys ’25, page 1279–1297. ACM, March 2025.
\newblock \doi{10.1145/3689031.3696075}.
\newblock URL \url{http://dx.doi.org/10.1145/3689031.3696075}.

\bibitem[Shi et~al.(2026)Shi, Ma, Liu, Zhao, Hwang, and Li]{shi2026intrinsicentropycontextlength}
Jingzhe Shi, Qinwei Ma, Hongyi Liu, Hang Zhao, Jeng-Neng Hwang, and Lei Li.
\newblock Intrinsic entropy of context length scaling in llms, 2026.
\newblock URL \url{https://arxiv.org/abs/2502.01481}.

\bibitem[Snell et~al.(2024)Snell, Lee, Xu, and Kumar]{snell2024scalingllmtesttimecompute}
Charlie Snell, Jaehoon Lee, Kelvin Xu, and Aviral Kumar.
\newblock Scaling llm test-time compute optimally can be more effective than scaling model parameters, 2024.
\newblock URL \url{https://arxiv.org/abs/2408.03314}.

\bibitem[Tso et~al.(2026)Tso, Schmittou, Huynh, and Hutchins]{tso2026constraintbenchbenchmarkingllmconstraint}
Joseph Tso, Preston Schmittou, Quan Huynh, and Jibran Hutchins.
\newblock Constraintbench: Benchmarking llm constraint reasoning on direct optimization, 2026.
\newblock URL \url{https://arxiv.org/abs/2602.22465}.

\bibitem[Wang et~al.(2024)Wang, Zhu, Han, Lin, Lin, Sun, and Ding]{wang2024optibench}
Zhuohan Wang, Ziwei Zhu, Yizhou Han, Yufeng Lin, Zhihang Lin, Ruoyu Sun, and Tian Ding.
\newblock Optibench: Benchmarking large language models in optimization modeling with equivalence-detection evaluation, 2024.
\newblock URL \url{https://openreview.net/forum?id=KD9F5Ap878}.

\bibitem[Xiao et~al.(2024)Xiao, Zhang, Wu, Xu, Wang, Han, Fu, Zhong, Zeng, Song, and Chen]{xiao2024chainofexperts}
Ziyang Xiao, Dongxiang Zhang, Yangjun Wu, Lilin Xu, Yuan~Jessica Wang, Xiongwei Han, Xiaojin Fu, Tao Zhong, Jia Zeng, Mingli Song, and Gang Chen.
\newblock Chain-of-experts: When {LLM}s meet complex operations research problems.
\newblock In \emph{The Twelfth International Conference on Learning Representations}, 2024.
\newblock URL \url{https://openreview.net/forum?id=HobyL1B9CZ}.

\bibitem[Xiao et~al.(2025)Xiao, Xie, Xu, Guan, Zhu, Han, Fu, Yu, Wu, Shi, Kang, Duan, Zhong, Yuan, Zeng, Wang, Chen, and Zhang]{xiao2025surveyoptimizationmodelingmeets}
Ziyang Xiao, Jingrong Xie, Lilin Xu, Shisi Guan, Jingyan Zhu, Xiongwei Han, Xiaojin Fu, WingYin Yu, Han Wu, Wei Shi, Qingcan Kang, Jiahui Duan, Tao Zhong, Mingxuan Yuan, Jia Zeng, Yuan Wang, Gang Chen, and Dongxiang Zhang.
\newblock A survey of optimization modeling meets llms: Progress and future directions, 2025.
\newblock URL \url{https://arxiv.org/abs/2508.10047}.

\bibitem[Zhang et~al.(2026)Zhang, Kraska, and Khattab]{zhang2026recursivelanguagemodels}
Alex~L. Zhang, Tim Kraska, and Omar Khattab.
\newblock Recursive language models, 2026.
\newblock URL \url{https://arxiv.org/abs/2512.24601}.

\bibitem[Zhang et~al.(2025)Zhang, Luo, Yang, Soong, and Yuen]{zhang2025orllmagentautomatingmodelingsolving}
Bowen Zhang, Pengcheng Luo, Genke Yang, Boon-Hee Soong, and Chau Yuen.
\newblock Or-llm-agent: Automating modeling and solving of operations research optimization problems with reasoning llm, 2025.
\newblock URL \url{https://arxiv.org/abs/2503.10009}.

\bibitem[Zheng et~al.(2024)Zheng, Zhang, Zhang, Ye, Luo, Feng, and Ma]{zheng2024llamafactory}
Yaowei Zheng, Richong Zhang, Junhao Zhang, Yanhan Ye, Zheyan Luo, Zhangchi Feng, and Yongqiang Ma.
\newblock Llamafactory: Unified efficient fine-tuning of 100+ language models.
\newblock In \emph{Proceedings of the 62nd Annual Meeting of the Association for Computational Linguistics (Volume 3: System Demonstrations)}, Bangkok, Thailand, 2024. Association for Computational Linguistics.
\newblock URL \url{http://arxiv.org/abs/2403.13372}.

\bibitem[Zhou et~al.(2025)Zhou, Liu, Chen, Tian, and Chen]{zhou2025gsminfinitellmsbehaveinfinitely}
Yang Zhou, Hongyi Liu, Zhuoming Chen, Yuandong Tian, and Beidi Chen.
\newblock Gsm-infinite: How do your llms behave over infinitely increasing context length and reasoning complexity?, 2025.
\newblock URL \url{https://arxiv.org/abs/2502.05252}.

\end{thebibliography}

\newpage
\appendix

\section{Dataset Curation Details}
\label{app:curation}

This appendix provides additional detail on the generation pipeline described in Section~\ref{subsec:pipeline}. Algorithm~\ref{alg:pipeline} gives the pseudocode.

\paragraph{Problem category details.}
Each template-based category requires a distinct optimization model:
\begin{itemize}[nosep,leftmargin=*]
    \item \textbf{Transportation}: Bipartite supply-demand LP.
    \item \textbf{Disaster Response}: Multi-period MILP with vehicle routing, supply shortages, and route security.
    \item \textbf{JSSP}: Job-shop scheduling with machine assignments and precedence constraints.
    \item \textbf{VRPTW}: Vehicle routing with time windows, capacity, and subtour elimination.
    \item \textbf{RCPSP}: Multi-mode project scheduling with time lags, budget, and deadlines.
    \item \textbf{Facility Location}: Deriving cost matrices from Euclidean distances (MILP).
    \item \textbf{Power Transmission}: Deriving quadratic power loss from Ohm's law (MIQP).
    \item \textbf{Queuing/Staffing}: Using Erlang-C formulas for service levels (nonlinear).
    \item \textbf{Stochastic Transportation}: Two-stage MILP with SAA and chance constraints.
    \item \textbf{Multi-Objective Transportation}: Bi-objective (cost + emissions) with fixed charges, MOQ, and supplier cardinality.
    \item \textbf{Modified Facility Location}: \textit{Facility location} with additional operational constraints.
\end{itemize}

\begin{algorithm}[h]
\caption{Text2Opt-Bench Generation Pipeline}
\label{alg:pipeline}
\begin{algorithmic}[1]
\State \textbf{Input:} Problem Type $T$, Dimensions $n, m$
\State $\mathcal{D}_{\text{struct}} \gets \text{GenerateWorldState}(T, n, m)$ \Comment{Domain-specific parameters}
\State $x^*, z^* \gets \text{SolverVerify}(\mathcal{D}_{\text{struct}})$ \Comment{Gurobi ground truth}
\If{Direct Translation (small scale)}
    \State $\mathcal{T} \gets \text{LLM}(\text{Prompt}_{\text{Direct}}, \mathcal{D}_{\text{struct}})$ \Comment{Full data in narrative}
\Else \Comment{Template-Based (large scale)}
    \State $\mathcal{T}_{\text{tmpl}} \gets \text{LLM}(\text{Prompt}_{\text{Template}}, \text{Schema}(\mathcal{D}_{\text{struct}}))$ \Comment{Structure only}
    \State $\mathcal{T} \gets \text{InsertData}(\mathcal{T}_{\text{tmpl}}, \mathcal{D}_{\text{struct}})$ \Comment{Fill placeholders}
\EndIf
\State \textbf{Output:} $(\mathcal{T}, \mathcal{D}_{\text{struct}}, x^*, z^*)$
\end{algorithmic}
\end{algorithm}

\subsection{Direct Translation: Mathematical Construction}
\label{app:direct_translation}

For resource allocation problems, we generate a linear programming problem in standard form:
\begin{equation}
    \begin{aligned}
    & \text{minimize} && c^T x \\
    & \text{subject to} && Ax \gtreqless b \\
    & && x \geq 0
    \end{aligned}
\end{equation}
To ensure control over the problem's characteristics, we use an \textit{anchor solution}:
\begin{enumerate}
    \item \textbf{Matrix Construction ($A$):} We initialize $A \in \mathbb{R}^{m \times n}$ with random values and apply a sparsity mask to simulate real-world interactions.
    \item \textbf{Anchor Solution ($x_{\text{anchor}}$):} We sample a feasible solution $x_{\text{anchor}} \geq 0$.
    \item \textbf{RHS Derivation ($b$):} The vector $b$ is derived via $b_i = (A x_{\text{anchor}})_i + s_i$, ensuring feasibility by construction.
\end{enumerate}

The structured representation is then passed to an LLM (GPT-5) with a prompt, and all numerical coefficients from $A$, $b$, and $c$ are put into a text description. An example of this process is in \S~\ref{app:data_embedding}.

\begin{algorithm}[h]
\caption{Direct Translation Dataset Generation}
\begin{algorithmic}[1]
\State \textbf{Input:} Dimensions $n, m$, Sparsity $S$
\State \Comment{\textbf{Phase 1: Construction (Guaranteed Feasibility)}}
\State $A \gets \text{RandomMatrix}(m, n, \text{sparsity}=S)$
\State $x_{\text{anchor}} \gets \text{RandomVector}(n, \min=0)$
\State $s \gets \text{RandomVector}(m, \min=0.5)$
\State $b \gets A x_{\text{anchor}} \pm s$ \Comment{Constructs $b$ s.t.\ $x_{\text{anchor}}$ is feasible}
\State $c \gets \text{RandomVector}(n)$
\State \Comment{\textbf{Phase 2: Verification (Ensured Optimality)}}
\State $x^*, z^*, \text{status} \gets \text{GurobiSolve}(A, b, c)$
\If{$\text{status} \neq \text{OPTIMAL}$}
    \State \textbf{return} \text{Retry} \Comment{Reject unbounded/infeasible}
\EndIf
\State $\mathcal{D}_{\text{struct}} \gets \{A, b, c, \text{senses}, \text{types}\}$
\State $\mathcal{T}_{\text{text}} \gets \text{LLM}(\text{SystemPrompt}, \mathcal{D}_{\text{struct}})$
\State \textbf{Output:} Pair $(\mathcal{T}_{\text{text}}, \mathcal{D}_{\text{struct}})$
\end{algorithmic}
\end{algorithm}

\subsection{Template-Based Generation: Full Pipeline}
\label{app:template_pipeline}

For structured problems (100+ variables), direct translation becomes impractical. We initially explored \textbf{hierarchical decomposition} via a block-diagonal structure ($A = \text{diag}(A_1, \dots, A_k)$), which decouples into $k$ independent sub-problems with $Z^* = \sum_{i=1}^k Z_i^*$. However, we discarded this approach due to three critical bottlenecks: (1) \textbf{context explosion}: merged narrative is too long; (2) \textbf{semantic fragmentation}: lack of global coherence in disjointed narratives; and (3) \textbf{topological inflexibility}: the method could not accommodate complex linking constraints.

To resolve this, we developed the \textbf{template-based pipeline}:

\paragraph{1. Structured Parameter Generation.}
Instead of a generic matrix $A$, we generate domain-specific parameters. For example, when generating facility location problems:
\begin{itemize}[noitemsep]
    \item Coordinates for $N$ facilities and $M$ customers.
    \item Fixed costs $f_i$, capacities $s_i$, demands $d_j$, and transport rates $r$.
\end{itemize}

Note: The transport cost matrix is not provided directly; the model must compute it from coordinates via Euclidean distance.

\paragraph{2. Template Generation via LLM.}
The LLM generates a template `` business memo'' describing the logic of the problem but \textit{excluding} numerical data.  Placeholders such as \texttt{\{CUSTOMER\_DEMANDS\}} are forced to be included.

\paragraph{3. Deterministic Data Insertion.}
The pipeline programmatically replaces placeholders with formatted generated data, decoupling linguistic complexity from numerical complexity.

\subsection{Data Embedding Example}
\label{app:data_embedding}

\begin{tcolorbox}[
    colback=white,
    colframe=black!75,
    title=\textbf{Example: Data Embedding Transformation},
    fonttitle=\bfseries,
    boxrule=0.8pt
]
\small
\textit{We sample a specific variable and constraint to demonstrate the mapping from structured parameters to natural language narrative.}

\medskip

\textbf{1. Variable Embedding} \\
\textbf{Input (Structured):}
\begin{itemize}
    \setlength\itemsep{0em}
    \item \texttt{Var\_1}: Type \texttt{Integer}, Obj Coeff $c_1 = 6.86$
    \item \texttt{Interaction}: Consumes $0.8$ of Resource $C_0$
\end{itemize}
\textbf{Output (Narrative):} \\
``\textbf{On-Site Retrofit Packages}: Each completed package adds \textbf{6.86} in contribution. Each package uses \textbf{0.8 units} from our environmental emissions allowance...''

\medskip
\hrule
\medskip

\textbf{2. Constraint Embedding} \\
\textbf{Input (Structured):}
\begin{itemize}
    \setlength\itemsep{0em}
    \item \texttt{Constraint C0}: Sense $\le$, RHS $b_0 = 8.25$
\end{itemize}
\textbf{Output (Narrative):} \\
``\textbf{Environmental Emissions Allowance}: Total available is \textbf{8.25} allowance units and cannot be exceeded.''
\end{tcolorbox}

\subsection{Evaluation Validity: False-Positive Prevention}
\label{app:eval_validity}

Our evaluation pipeline is designed to minimize false positives at two levels.

\paragraph{Feasibility by construction.} If infeasible problems were included, a model producing a wrong formulation would frequently also yield an infeasible result, creating a false positive under code-result evaluation. Restricting to feasible instances ensures that any infeasible output is unambiguously incorrect.

\paragraph{Objective fingerprinting.} The remaining false-positive risk is a structurally different formulation that coincidentally matches the gold objective. In our pipeline, all instances use randomly generated continuous coefficients with wide ranges (e.g., costs from $[5,30]$, demands from $[10,100]$), making the optimal objective an effective fingerprint: coincidental agreement to $10^{-4}$ relative tolerance is negligible. We avoid supplementing objective matching with variable/constraint count checks, as correct formulations can legitimately differ in these counts due to auxiliary variables (e.g., $t = \max(x,y)$), constraint decomposition, or alternative modeling strategies (e.g., Miller--Tucker--Zemlin vs.\ lazy subtour elimination in VRPTW).

\subsection{Robustness to the Description Generator}
\label{app:description_robustness}

\paragraph{Self-contamination.} GPT-5 is used both to generate benchmark instances and as an evaluated model, raising a potential self-contamination concern. For template-based categories, this concern is structurally precluded: GPT-5 generates only the prose template (natural language structure), while all numerical data is inserted deterministically by scripts.

For resource allocation (direct translation), GPT-5 generates the full problem description including numerical coefficients. However, two observations argue against contamination: (1)~GPT-5 (87.9\%) is \emph{outperformed} by Claude Opus~4.6 (89.9\%), which is not included in generation; (2)~Table~\ref{tab:failure_modes} shows that GPT-5's failures are predominantly binding errors, similar to other models — memorization would primarily aid coefficient recall, yet GPT-5 shows no such advantage.

\paragraph{Author-model robustness.} All problem descriptions in Text2Opt-Bench are authored via GPT-5. To rule out artifacts resulting from the use of a single author, we regenerated all prose with Claude Opus 4.6 (here, instance data and gold solutions were held fixed) and re-evaluated three models on both resource allocation (RA) and the template categories (Tmpl), running standard inference twice per problem (n=50 problems $\times$ 2 runs = 100 paired observations per cell). Five of six cells are statistically not significant (using a paired McNemar p $\geq 0.37$). The one marginally significant cell (Claude-Opus on RA, $-7$pp, p=0.023) goes \emph{against} its own family --- Claude's own prose lowers Claude-Opus's score --- which strengthens the no-self-favoring claim.

\begin{table}[h]
\centering
\caption{Description-author robustness. Mean accuracy across 2 runs per (model, source, problem) cell; Wilson 95\% CIs and paired McNemar use all (problem, run) pairs (n=100 per row).}
\label{tab:description_robustness}
\footnotesize
\begin{tabular}{llcccccc}
\toprule
Model & Dataset & GPT-5 authored (95\% CI) & Claude-authored (95\% CI) & $\Delta$ & b/c & McNemar p \\
\midrule
GPT-5-Nano  & RA   & 55.0 [45.2, 64.4]  & 58.0 [48.2, 67.2]  & +3pp & 14/11 & p=0.69 n.s. \\
GPT-5       & RA   & 96.0 [90.2, 98.4]  & 99.0 [94.6, 99.8]  & +3pp & 4/1   & p=0.37 n.s. \\
Claude-Opus & RA   & 97.0 [91.5, 99.0]  & 90.0 [82.6, 94.5]  & $-7$pp & 0/7 & p=0.023 * \\
GPT-5-Nano  & Tmpl & 67.0 [57.3, 75.4]  & 71.0 [61.5, 79.0]  & +4pp & 14/10 & p=0.54 n.s. \\
GPT-5       & Tmpl & 89.0 [81.4, 93.7]  & 86.0 [77.9, 91.5]  & $-3$pp & 3/6 & p=0.50 n.s. \\
Claude-Opus & Tmpl & 92.0 [85.0, 95.9]  & 91.0 [83.8, 95.2]  & $-1$pp & 5/6 & p=1.00 n.s. \\
\bottomrule
\end{tabular}
\end{table}

\section{Failure Mode Analysis}
\label{app:failure_modes}

To validate that resource allocation is a binding-dominated task, we classify every failure across 9 models by checking the Gurobi model structure of the generated code.

\paragraph{Classification.}
For each failed solution, we extract the number of decision variables (\texttt{NumVars}) and constraints (\texttt{NumConstrs}) from the Gurobi model object and compare against the gold solution:
\begin{itemize}[nosep,leftmargin=*]
    \item \textbf{Binding error}: correct \texttt{NumVars} and \texttt{NumConstrs} but wrong objective value---the model understood the formulation but mis-transcribed coefficients.
    \item \textbf{Modeling error}: incorrect \texttt{NumVars} or \texttt{NumConstrs}---the model produced a structurally different formulation. 
    \item \textbf{Execution error}: the generated code fails to execute (syntax errors, runtime exceptions).
\end{itemize}

\paragraph{Results.}
Table~\ref{tab:failure_modes} reports the failure breakdown across the 248 resource allocation eval-subset instances. For every model except Qwen2.5-7B, failures are overwhelmingly due to incorrect coefficient transcription (binding errors), which account for 60.4–92.3\% of the total, while modeling errors remain negligible at 0–3\%. Qwen2.5-7B is the sole exception, with a 14.9\% modeling-error rate that reflects its weak formulation ability (discussed below).

\begin{table}[h]
\centering
\caption{Failure mode breakdown on resource allocation (248 eval-subset instances). Percentages are of total failures per model. Binding errors dominate for all models.}
\label{tab:failure_modes}
\begin{tabular}{lrrrrr}
\toprule
Model & Pass\% & Fail & Exec\% & Model\% & Bind\% \\
\midrule
Claude Opus 4.6   & 89.9 & 25  & 36.0 & 0.0 & 64.0 \\
GPT-5             & 87.9 & 30  & 16.7 & 0.0 & 83.3 \\
Claude Sonnet 4.6 & 84.7 & 38  & 15.8 & 0.0 & 84.2 \\
DeepSeek-R1       & 80.6 & 48  & 37.5 & 2.1 & 60.4 \\
o4-mini           & 80.2 & 49  & 14.3 & 0.0 & 75.5 \\
DeepSeek-V3.2     & 79.0 & 52  &  5.8 & 1.9 & 92.3 \\
Llama3.3-70B     & 49.6 & 125 & 25.6 & 3.2 & 71.2 \\
GPT-5-Nano        & 49.2 & 126 & 21.4 & 2.4 & 76.2 \\
Qwen2.5-7B        & 13.3 & 215 & 22.3 & 14.9 & 62.8 \\
\bottomrule
\end{tabular}
\end{table}

\paragraph{Case study.}
A representative binding error from GPT-5 involves a problem with 12 variables and 13 constraints. The generated code perfectly reproduces all variable bounds, the objective function, and 12 of the 13 constraints. However, one constraint (``On-Time Delivery Deviation'') includes an extra coefficient (\texttt{3.16\,*\,x1}) that leaked from a different constraint (``Safety Risk Index''). The model understood the structure but misplaced a single coefficient, shifting the optimal objective from 581.41 to 580.06.

\paragraph{Sensitivity to evaluation tolerance.}
Many binding errors produce near-optimal solutions: for Claude Opus 4.6, 100\% of binding failures have relative objective error below 5\%; for GPT-5, 87\% are within 5\% (median 1.5\%). Under a relaxed 5\% tolerance, these would all pass---but this inflates scores without changing the relative model ranking or the binding-vs-modeling conclusion.

\paragraph{Note on Qwen2.5-7B.}
Qwen2.5-7B has the highest modeling error rate (14.9\%) and execution error rate (22.3\%) of any model, reflecting insufficient code generation and formulation capability at this scale. Its remaining failures are still predominantly binding errors (62.8\%), consistent with the overall pattern.

\paragraph{Isomorphism validation of passing solutions.}
A potential concern is that passing solutions achieve the correct objective with a structurally different formulation (``correct for the wrong reasons''). We validate this by extracting the full Gurobi model (objective, bounds, constraint matrix, RHS, senses) from both gold and generated code, comparing them under a standardized canonical form. Across all passing solutions from the 9 models, 90.5\% are provably isomorphic. A mutual feasibility check confirms another 2.6\% are algebraically equivalent reformulations. The final 6.8\% have different feasible regions but identical optima; manual inspection reveals these are valid variable permutations and algebraic rewrites unresolved by our canonical sort (e.g., $-5.4x \geq -9.34$ rewritten as $x \leq 1.73$). We found no instances of genuinely different formulations coincidentally matching the objective.

\paragraph{Why this analysis does not extend to template problems.}
As described in \S\ref{app:eval_validity}, the analysis is less informative for template problems. An alternative approach---automated error classification via LLM---is possible in principle but unreliable at scale due to binding limitation. For template problems, BIND provides a cleaner diagnostic: categories where BIND recovers most failures are binding-limited, while categories where BIND provides no gain are modeling-limited (\S\ref{subsubsec:inference}).

\subsection{Failure Modes of an Untrained Binding Stage (No-Oracle 2-Stage Pipeline)}
\label{app:two_stage_failures}
To test if the binding/modeling decomposition works without a trained specialist, we evaluate a 2-stage inference pipeline using a single frontier model. Stage~1 extracts the instance data JSON from the prose; Stage~2 writes Gurobi code to read that JSON. Evaluated on 100 problems (50 queuing/staffing + 50 stochastic transportation) without an oracle, Stage-1 successfully parsed the JSON in 85/100 instances. Table~\ref{tab:two_stage} compares this setup against vanilla pass@1 and BIND with oracle JSON.

\begin{table}[h]
\centering
\caption{2-stage no-oracle pipeline (gpt-5-nano, n=100, 50 qs + 50 st). Paired McNemar reveals that splitting binding from modeling at inference time \emph{hurts} accuracy: an untrained frontier model is too noisy at binding. The trained binding specialist of \S\ref{subsubsec:training} closes this gap by learning canonical schemas and complete-field extraction.}
\label{tab:two_stage}
\footnotesize
\begin{tabular}{lcccc}
\toprule
Condition & Overall (95\% CI) & qs (n=50) & st (n=50) & paired McNemar vs Vanilla \\
\midrule
Vanilla pass@1            & 47.0 [37.5, 56.7] & 58.0 [44.2, 70.6] & 36.0 [24.1, 49.9] & --- \\
BIND oracle (1-pass)      & 78.0 [68.9, 85.0] & 94.0 [83.8, 97.9] & 62.0 [48.2, 74.1] & b/c=40/9, p=1.8e-5 *** \\
2-stage no-oracle         & 31.0 [22.8, 40.6] & 40.0 [27.6, 53.8] & 22.0 [12.8, 35.2] & b/c=6/22, p=0.005 ** \\
\bottomrule
\end{tabular}
\end{table}

\paragraph{Failure modes of the no-oracle stage~1.} Our manual inspection of the cases where BIND-oracle succeeded but 2-stage failed revealed four characteristic failure modes:
\begin{enumerate}[noitemsep,topsep=2pt,leftmargin=1.5em]
\item \textbf{Field renaming.} Stage~1 invents its own naming convention (e.g.\ \texttt{n\_stations} $\to$ \texttt{number\_of\_stations}, \texttt{recourse\_costs} $\to$ \texttt{emergency\_shipping\_cost\_matrix}).
\item \textbf{Wrong scale or unit.} Reading ``90\% confidence'' as the integer 90 instead of the probability 0.9, which is then fed to a chance constraint, producing wrong solutions.
\item \textbf{Sentinel-encoding errors.} Emitting \texttt{None} where the gold value is \texttt{0.0} (representing ``machine doesn't process job''); the downstream code then crashes on arithmetic.
\item \textbf{Missing critical fields.} Omitting entire fields needed by the formulation (e.g.\ \texttt{max\_violations\_per\_dest}, \texttt{recourse\_capacities} for chance-constrained transportation), which the downstream stage cannot recover.
\end{enumerate}

\section{Prompting Ablation}
\label{app:prompting}

To investigate whether advanced prompting improves performance, we conducted an ablation study with GPT-5-Nano on resource allocation problem subset ($n$=248). Table~\ref{tab:prompting_results} shows that no prompting strategy improves over the base prompt noticeably. Every variant---extra reasoning, explicit warnings, additional focus requirements, second-pass refinement, and one-shot examples---either performs similarly or degrades accuracy, with one-shot examples dropping accuracy to 44.4\%. We attribute this to the effective context limit: the bottleneck is not instruction quality but the model's capacity to faithfully process dense numerical specifications, which prompting alone cannot address.

\begin{table}[h]
\caption{Prompting strategy ablation on GPT-5-Nano (resource allocation, $n$=248).}
\label{tab:prompting_results}
\begin{center}
\begin{tabular}{lc}
\toprule
\textbf{Prompting Strategy} & \textbf{Accuracy (\%)} \\
\midrule
Base Prompt (with Template) & 49.2 \\
Base + Extra Reasoning & 50.0 \\
Base + Explicit Warnings & 45.7 \\
Base + Additional Focus Requirements & 47.6 \\
Base + Second Pass Refinement & 50.0 \\
One-Shot Example & 44.4 \\
\bottomrule
\end{tabular}
\end{center}
\end{table}

\subsection{Transcription vs Prompt-Length Ablation (BIND mechanism)}
\label{app:bind_ablation_padding}

To isolate whether BIND's gain comes from removing transcription or simply from a shorter/cleaner prompt, we compare three single-pass conditions (no refinement) on three binding-limited categories (queuing/staffing, stochastic transportation, RCPSP; 50 problems each, n=150 total): (i) \emph{Standard Inference} --- data inline in the prompt (model transcribes); (ii) \emph{BIND} --- data offloaded to a JSON file; (iii) \emph{Padded BIND} --- BIND plus neutral, clearly-irrelevant filler text (e.g.\ \emph{``Standard operating procedures, reporting cadences, governance reviews\dots''}) so the prompt's length matches Standard Inference. Tokenization uses GPT-5-Nano's tokenizer as a fixed proxy across model families.

\begin{table}[h]
\centering
\caption{Wilson 95\% CIs (n=150) and continuity-corrected paired McNemar p-values. For every model, BIND $\gg$ Standard is highly significant (p $\leq 1.6{\times}10^{-6}$); padding BIND back to baseline length does not significantly change accuracy (p $\geq 0.27$), so the gain comes from removing transcription, not from a shorter prompt.}
\label{tab:bind_padding}
\footnotesize
\begin{tabular}{lccccc}
\toprule
Model & Standard (95\% CI) & BIND (95\% CI) & Padded BIND (95\% CI) & BIND vs Std. & Padded vs BIND \\
\midrule
GPT-5-Nano      & 38.0 [30.6, 46.0] & 65.3 [57.4, 72.5] & 64.0 [56.1, 71.2] & p=1.5e-6 *** & p=0.89 n.s. \\
DeepSeek-V3.2   & 48.7 [40.8, 56.6] & 69.3 [61.5, 76.2] & 74.7 [67.2, 81.0] & p=3.8e-5 *** & p=0.27 n.s. \\
Claude Opus 4.6 & 83.3 [76.6, 88.4] & 100.0 [97.5, 100.0] & 98.7 [95.3, 99.6] & p=1.6e-6 *** & p=0.48 n.s. \\
\bottomrule
\end{tabular}
\end{table}

\section{RULER Binding Task Implementation Details}
\label{app:ruler}

We adapt the RULER long-context benchmark~\citep{hsieh2024rulerwhatsrealcontext} with modifications designed to prevent ceiling effects at larger model sizes. The specifications are listed below.

\paragraph{Task descriptions.}
\begin{itemize}[nosep,leftmargin=*]
    \item \textbf{Single-key retrieval} (\textsc{niah\_single}): retrieve one UUID value associated with a specific key embedded in the haystack, analogous to reading a single parameter.
    \item \textbf{Multi-key retrieval} (\textsc{niah\_multikey}): retrieve UUID values for $N$ distinct keys simultaneously, analogous to binding all coefficients in a constraint.
    \item \textbf{Multi-value retrieval} (\textsc{niah\_multivalue}): recall all $N$ UUID values associated with a single key that appears multiple times, analogous to reading an entire data column.
    \item \textbf{Aggregation}: locate records scattered across multiple categories and compute a count for a target category, the operation closest to assembling an objective function from distributed data.
\end{itemize}

\paragraph{Task generation.} Each task generates synthetic prompts at six target context lengths: 1K, 2K, 4K, 8K, 16K, and 32K tokens (measured via the Qwen-2.5 tokenizer). Prompts consist of a \emph{haystack} of expository prose paragraphs from Paul Graham essays, following the original RULER implementation \cite{hsieh2024rulerwhatsrealcontext}, with task-specific \emph{needles} (key-value pairs or records) inserted at uniformly random positions. A question requiring extraction of the embedded information is appended at the end. We generate 200 samples per task per context length (1{,}200 per task, 4{,}800 total across four tasks).

\paragraph{Hardening against easy retrieval.} The original RULER tasks use unique, easily distinguishable keys. We introduce three forms of increased difficulty that scale with context length:

\begin{itemize}[nosep,leftmargin=*]
    \item \textbf{Distractor needles with confusable names.} For single-key and multi-value tasks, distractor keys are generated by substituting one character in the target key name (e.g., \texttt{special\_item\_abcde} vs.\ \texttt{special\_item\_abcdf}). The number of distractors scales as $\max(3, L/1024)$ where $L$ is the target token length.
    \item \textbf{Scaled binding complexity.} For multi-key and multi-value tasks, the number of target values to retrieve scales as $\max(2, L/2048)$, with $3\times$ distractors per real key. At 32K tokens, the model must retrieve 15 values amidst 45 distractors.
    \item \textbf{Category-based aggregation.} The aggregation task scatters records across $\max(3, L/4096 + 2)$ categories with $\max(3, L/2048)$ items each. The model must count or sum values for a single target category while ignoring all others.
\end{itemize}

\paragraph{Evaluation.} We evaluate six models from Qwen2.5-Instruct family (0.5B–32B) using vLLM~\citep{kwon2023efficientmemorymanagementlarge} with greedy decoding ($temperature=0$) and a maximum generation length of 128 tokens. We use \emph{strict exact-match} scoring for all tasks with no partial credit. This all-or-nothing criterion is motivated by optimization evaluation, where a single incorrect parameter yields an incorrect result. We did not include closed-source frontier models because these tasks are falsely flagged as jailbreaking attempts by the content filter.
\section{Full TTC and BIND Results}
\label{app:ttc_full}

\begin{table}[H]
\centering
\caption{Test-Time-Compute: Accuracy (\%) and Total Tokens (input + output, in K) per Problem across Models and Methods (550 Template Problems). Pass@5 represents the parallel upper bound. Repair@5 uses feedback from an oracle verifier (objective value and structure) each round. BIND is data offloading with the same oracle-guided refinement ($\le$3 rounds); its token cost is summed across all rounds. BIND matches the inline Repair@5 baseline at lower token cost because offloading the data shrinks the per-round input the loop re-sends. A single-pass ablation (no refinement) is in Table~\ref{tab:bind_ablation}.}
\label{tab:ttc_full}
\resizebox{\textwidth}{!}{%
\begin{tabular}{l|cc|cc|cc|cc|cc}
\toprule
 & \multicolumn{2}{c|}{Pass@1} & \multicolumn{2}{c|}{BIND} & \multicolumn{2}{c|}{Maj.\ Vote} & \multicolumn{2}{c|}{Pass@5} & \multicolumn{2}{c}{Repair@5\textsuperscript{$\dagger$}} \\
Model & Acc & Tok & Acc & Tok & Acc & Tok & Acc & Tok & Acc & Tok \\
\midrule
Claude Sonnet 4.6 & 87.6 & 4.4K & 96.2 & 3.4K & 89.6 & 22.0K & 97.8 & 22.0K & 98.4 & 6.5K \\
Claude Opus 4.6 & 86.7 & 4.4K & 98.7 & 3.3K & 88.4 & 21.8K & 96.0 & 21.8K & 98.7 & 6.3K \\
GPT-5 & 86.2 & 4.2K & 95.8 & 3.3K & 91.5 & 21.0K & 95.8 & 21.0K & 95.5 & 6.9K \\
o4-mini & 80.4 & 4.1K & 94.7 & 3.3K & 83.6 & 20.4K & 94.9 & 20.4K & 95.1 & 12.0K \\
DeepSeek-R1 & 72.9 & 3.8K & 84.5 & 4.4K & 78.4 & 18.8K & 93.6 & 18.8K & 92.4 & 7.1K \\
DeepSeek-V3.2 & 72.0 & 4.6K & 87.1 & 4.8K & 69.6 & 22.8K & 89.1 & 22.8K & 88.7 & 10.7K \\
GPT-5-Nano & 59.1 & 3.8K & 82.4 & 5.1K & 61.8 & 19.2K & 82.0 & 19.2K & 78.2 & 12.2K \\
\midrule
Llama3.3-70B & 35.1 & 4.0K & 46.0 & 7.7K & 33.3 & 20.1K & 51.6 & 20.1K & 50.7 & 19.5K \\
Qwen2.5-7B & 4.5 & 3.6K & 8.9 & 2.4K & 2.7 & 18.0K & 10.5 & 18.0K & 8.9 & 28.0K \\
\bottomrule
\end{tabular}}
\par\vspace{3pt}
\begin{minipage}{\textwidth}
\footnotesize \textsuperscript{$\dagger$}Repair@5 uses oracle feedback (ground-truth objective + structure) each round. BIND applies the same refinement with data offloaded, and its token cost is summed across rounds.
\end{minipage}
\end{table}

\begin{table}[H]
\caption{BIND per-category accuracy (\%, n=50 per category). $\Delta$ = improvement over default (data in prompt). BIND helps most on data-heavy categories for capable models, but cannot fix modeling gaps in weaker models. A single-pass ablation (data offloading only, no refinement) is in Table~\ref{tab:bind_ablation}.}
\label{tab:bind}
\begin{center}
\small
\setlength{\tabcolsep}{2pt}
\resizebox{\textwidth}{!}{%
\begin{tabular}{l|cc|cc|cc|cc|cc|cc|cc|cc|cc}
\toprule
 & \multicolumn{2}{c|}{\textbf{GPT-5}} & \multicolumn{2}{c|}{\textbf{Opus}} & \multicolumn{2}{c|}{\textbf{Sonnet}} & \multicolumn{2}{c|}{\textbf{o4-mini}} & \multicolumn{2}{c|}{\textbf{DS-R1}} & \multicolumn{2}{c|}{\textbf{DS-V3.2}} & \multicolumn{2}{c|}{\textbf{Nano}} & \multicolumn{2}{c|}{\textbf{Llama}} & \multicolumn{2}{c}{\textbf{Qwen}} \\
 & B & $\Delta$ & B & $\Delta$ & B & $\Delta$ & B & $\Delta$ & B & $\Delta$ & B & $\Delta$ & B & $\Delta$ & B & $\Delta$ & B & $\Delta$ \\
\midrule
Transp. & 100 & 0 & 100 & \tg{+2} & 100 & 0 & 100 & 0 & 94 & \tr{$-$6} & 100 & 0 & 100 & 0 & 100 & \tg{+12} & 82 & \tg{+44} \\
Disaster & 100 & \tg{+14} & 100 & \tg{+4} & 100 & \tg{+4} & 100 & \tg{+6} & 66 & \tr{$-$12} & 88 & \tr{$-$2} & 84 & \tg{+22} & 60 & \tg{+30} & 0 & 0 \\
JSSP & 100 & \tg{+10} & 100 & 0 & 100 & \tg{+2} & 100 & \tg{+4} & 100 & \tg{+4} & 96 & 0 & 100 & \tg{+18} & 0 & 0 & 0 & 0 \\
VRPTW & 96 & \tg{+26} & 100 & \tg{+62} & 92 & \tg{+42} & 82 & \tg{+48} & 66 & \tg{+32} & 48 & \tg{+26} & 36 & \tg{+34} & 2 & \tg{+2} & 0 & 0 \\
RCPSP & 100 & \tg{+12} & 100 & \tg{+4} & 100 & 0 & 98 & \tg{+16} & 94 & \tg{+60} & 94 & \tg{+32} & 68 & \tg{+42} & 2 & \tg{+2} & 0 & 0 \\
Fac.\ Loc. & 100 & 0 & 100 & 0 & 100 & \tg{+2} & 100 & \tg{+2} & 98 & \tg{+4} & 100 & \tg{+2} & 100 & \tg{+10} & 98 & 0 & 12 & \tg{+6} \\
Power T. & 80 & \tr{$-$18} & 86 & \tr{$-$2} & 66 & \tg{+2} & 82 & \tg{+12} & 66 & \tg{+2} & 72 & \tg{+18} & 68 & \tg{+18} & 12 & \tr{$-$4} & 0 & 0 \\
Queue/St. & 98 & \tg{+18} & 100 & \tg{+8} & 100 & \tg{+2} & 100 & \tg{+24} & 100 & \tg{+30} & 94 & \tg{+28} & 98 & \tg{+42} & 30 & \tg{+20} & 0 & 0 \\
Stoch.\ T. & 96 & \tg{+26} & 100 & \tg{+38} & 100 & \tg{+38} & 100 & \tg{+34} & 66 & \tg{+6} & 76 & \tg{+58} & 88 & \tg{+56} & 22 & \tg{+16} & 0 & 0 \\
M-Obj T. & 84 & \tg{+14} & 100 & \tg{+12} & 100 & \tg{+2} & 80 & \tg{+12} & 86 & \tg{+10} & 90 & \tg{+4} & 72 & \tg{+12} & 86 & \tg{+44} & 2 & \tr{$-$2} \\
Mod.\ FL & 100 & \tg{+4} & 100 & \tg{+4} & 100 & 0 & 100 & 0 & 94 & \tr{$-$2} & 100 & 0 & 92 & \tg{+2} & 94 & \tr{$-$2} & 2 & 0 \\
\midrule
\textbf{Avg} & \textbf{95.8} & \tg{+9.6} & \textbf{98.7} & \tg{+12.0} & \textbf{96.2} & \tg{+8.6} & \textbf{94.7} & \tg{+14.4} & \textbf{84.5} & \tg{+11.6} & \textbf{87.1} & \tg{+15.1} & \textbf{82.4} & \tg{+23.3} & \textbf{46.0} & \tg{+10.9} & \textbf{8.9} & \tg{+4.4} \\
\bottomrule
\end{tabular}}
\end{center}
\end{table}

\begin{table}[H]
\caption{\textbf{Single-pass ablation of BIND} (data offloading only; no refinement). Per-category accuracy (\%, n=50); $\Delta$ = change vs.\ default (data in prompt). This isolates how much of BIND's gain is pure transcription removal: a single offloaded pass already recovers $+12$ to $+27$pp on binding-limited categories for every model, but regresses on modeling-limited ones, where the formulation is the bottleneck and the gain in Table~\ref{tab:bind} comes from refinement. Qwen2.5-7B cannot model and stays near 0\% regardless.}
\label{tab:bind_ablation}
\begin{center}
\small
\setlength{\tabcolsep}{2pt}
\resizebox{\textwidth}{!}{%
\begin{tabular}{l|cc|cc|cc|cc|cc|cc|cc|cc|cc}
\toprule
 & \multicolumn{2}{c|}{\textbf{GPT-5}} & \multicolumn{2}{c|}{\textbf{Opus}} & \multicolumn{2}{c|}{\textbf{Sonnet}} & \multicolumn{2}{c|}{\textbf{o4-mini}} & \multicolumn{2}{c|}{\textbf{DS-R1}} & \multicolumn{2}{c|}{\textbf{DS-V3.2}} & \multicolumn{2}{c|}{\textbf{Nano}} & \multicolumn{2}{c|}{\textbf{Llama}} & \multicolumn{2}{c}{\textbf{Qwen}} \\
 & B & $\Delta$ & B & $\Delta$ & B & $\Delta$ & B & $\Delta$ & B & $\Delta$ & B & $\Delta$ & B & $\Delta$ & B & $\Delta$ & B & $\Delta$ \\
\midrule
Transp. & 100 & 0 & 100 & \tg{+2} & 100 & 0 & 100 & 0 & 48 & \tr{$-$52} & 100 & 0 & 100 & 0 & 100 & \tg{+12} & 82 & \tg{+44} \\
Disaster & 78 & \tr{$-$8} & 90 & \tr{$-$6} & 84 & \tr{$-$12} & 92 & \tr{$-$2} & 56 & \tr{$-$22} & 74 & \tr{$-$16} & 68 & \tg{+6} & 40 & \tg{+10} & 0 & 0 \\
JSSP & 96 & \tg{+6} & 100 & 0 & 100 & \tg{+2} & 96 & 0 & 94 & \tr{$-$2} & 92 & \tr{$-$4} & 88 & \tg{+6} & 0 & 0 & 0 & 0 \\
VRPTW & 66 & \tr{$-$4} & 8 & \tr{$-$30} & 62 & \tg{+12} & 48 & \tg{+14} & 26 & \tr{$-$8} & 12 & \tr{$-$10} & 10 & \tg{+8} & 0 & 0 & 0 & 0 \\
RCPSP & 100 & \tg{+12} & 100 & \tg{+4} & 96 & \tr{$-$4} & 90 & \tg{+8} & 68 & \tg{+34} & 88 & \tg{+26} & 40 & \tg{+14} & 0 & 0 & 0 & 0 \\
Fac.\ Loc. & 100 & 0 & 100 & 0 & 100 & \tg{+2} & 98 & 0 & 94 & 0 & 100 & \tg{+2} & 98 & \tg{+8} & 90 & \tr{$-$8} & 12 & \tg{+6} \\
Power T. & 40 & \tr{$-$58} & 54 & \tr{$-$34} & 44 & \tr{$-$20} & 66 & \tr{$-$4} & 44 & \tr{$-$20} & 44 & \tr{$-$10} & 46 & \tr{$-$4} & 6 & \tr{$-$10} & 0 & 0 \\
Queue/St. & 98 & \tg{+18} & 100 & \tg{+8} & 100 & \tg{+2} & 100 & \tg{+24} & 90 & \tg{+20} & 86 & \tg{+20} & 88 & \tg{+32} & 16 & \tg{+6} & 0 & 0 \\
Stoch.\ T. & 96 & \tg{+26} & 100 & \tg{+38} & 100 & \tg{+38} & 88 & \tg{+22} & 48 & \tr{$-$12} & 34 & \tg{+16} & 68 & \tg{+36} & 2 & \tr{$-$4} & 0 & 0 \\
M-Obj T. & 74 & \tg{+4} & 92 & \tg{+4} & 100 & \tg{+2} & 76 & \tg{+8} & 54 & \tr{$-$22} & 72 & \tr{$-$14} & 68 & \tg{+8} & 18 & \tr{$-$24} & 2 & \tr{$-$2} \\
Mod.\ FL & 96 & 0 & 100 & \tg{+4} & 100 & 0 & 88 & \tr{$-$12} & 90 & \tr{$-$6} & 96 & \tr{$-$4} & 86 & \tr{$-$4} & 90 & \tr{$-$6} & 2 & 0 \\
\midrule
\textbf{Avg} & \textbf{85.8} & \tr{$-$0.4} & \textbf{85.8} & \tr{$-$0.9} & \textbf{89.6} & \tg{+2.0} & \textbf{85.6} & \tg{+5.2} & \textbf{64.7} & \tr{$-$8.2} & \textbf{72.5} & \tg{+0.5} & \textbf{69.1} & \tg{+10.0} & \textbf{32.9} & \tr{$-$2.2} & \textbf{8.9} & \tg{+4.4} \\
\bottomrule
\end{tabular}}
\end{center}
\end{table}

\subsection{Statistical Significance for BIND vs pass@\textit{k} (n=550)}
\label{app:wilson_mcnemar}

We add Wilson 95\% confidence intervals and continuity-corrected paired McNemar tests for BIND vs pass@1 and BIND vs pass@5 (n=550 template problems; all three conditions share problem IDs, so the McNemar test is the appropriate paired comparison). Table~\ref{tab:wilson_mcnemar} reports the full per-model breakdown.

\begin{table}[h]
\centering
\caption{Wilson 95\% CIs (in brackets) and continuity-corrected paired McNemar p-values on n=550 template problems. BIND significantly exceeds pass@1 for every model (p ranging from $5.3{\times}10^{-4}$ for Qwen2.5-7B up to $1.1{\times}10^{-22}$ for GPT-5-Nano); BIND matches pass@5 (p $\geq 0.15$) for 7 of 9 models, at substantially lower token cost (Table~\ref{tab:ttc_full}).}
\label{tab:wilson_mcnemar}
\footnotesize
\begin{tabular}{lcccccc}
\toprule
Model & pass@1 (95\% CI) & BIND (95\% CI) & pass@5 (95\% CI) & BIND vs pass@1 & BIND vs pass@5 \\
\midrule
Claude Sonnet 4.6 & 87.6 [84.6, 90.1] & 96.2 [94.2, 97.5] & 97.8 [96.2, 98.7] & p=1.1e-7 *** & p=0.15 n.s. \\
Claude Opus 4.6   & 86.7 [83.6, 89.3] & 98.7 [97.4, 99.4] & 96.0 [94.0, 97.3] & p=3.7e-13 ***& p=0.01 ** \\
GPT-5             & 86.2 [83.0, 88.8] & 95.8 [93.8, 97.2] & 95.8 [93.8, 97.2] & p=1.9e-9 *** & p=0.86 n.s. \\
GPT-5-Nano        & 59.1 [54.9, 63.1] & 82.4 [79.0, 85.3] & 82.0 [78.6, 85.0] & p=1.1e-22 ***& p=0.92 n.s. \\
o4-mini           & 80.4 [76.8, 83.5] & 94.7 [92.5, 96.3] & 94.9 [92.7, 96.5] & p=8.4e-15 ***& p=1.00 n.s. \\
DeepSeek-R1       & 72.9 [69.0, 76.5] & 84.5 [81.3, 87.3] & 93.6 [91.3, 95.4] & p=4.6e-7 *** & p$<$0.01 *** \\
DeepSeek-V3.2     & 72.0 [68.1, 75.6] & 87.1 [84.0, 89.6] & 89.1 [86.2, 91.4] & p=7.8e-13 ***& p=0.24 n.s. \\
Llama-3.3-70B     & 35.1 [31.2, 39.2] & 46.0 [41.9, 50.2] & 51.6 [47.5, 55.8] & p=1.4e-8 *** & p$<$0.01 ** \\
Qwen2.5-7B        & 4.5 [3.1, 6.6]    & 8.9 [6.8, 11.6]   & 10.5 [8.2, 13.4]  & p=5.3e-4 *** & p=0.25 n.s. \\
\bottomrule
\end{tabular}
\end{table}

\subsection{Per-Category Limitation Regime}
\label{app:category_regimes}

For each category, we compare each model's performance with and without oracle data-offloading. We classify a category as \emph{binding-limited} if the majority of models gain from offloading, \emph{modeling-limited} if offloading shows no gain or a regression, and \emph{mixed} otherwise.

\begin{table}[h]
\centering
\caption{Per-category limitation regime. Most categories are binding-limited; a handful retain a modeling component. This refines the headline claim that ``binding is the primary bottleneck'' to the per-category nuance.}
\label{tab:category_regimes}
\footnotesize
\begin{tabular}{ll}
\toprule
Regime & Categories \\
\midrule
\textbf{Binding-limited}   & Transportation, Facility Loc., Mod. Fac. Loc., JSSP, RCPSP, Queuing/Staffing, Stochastic Transp. \\
\textbf{Mixed}             & Multi-Objective Transp., VRPTW, Disaster Response \\
\textbf{Modeling-limited}  & Power Transmission \\
\bottomrule
\end{tabular}
\end{table}

\subsection{Comparison Against Search-over-Formulations Baselines (SolverLLM/MCTS)}
\label{app:mcts_baseline}

We compare BIND against SolverLLM~\citep{li2025solverllmleveragingtesttimescaling}, an MCTS-based strategy that searches over candidate \emph{formulations} (modeling choices) to produce solver code. SolverLLM is run at 5 search iterations to match pass@5, on all 550 template problems. The oracle variant uses ground-truth objective values as the verifier; the deployable (no-oracle) variant uses the model's self-judgment.

\begin{table}[h]
\centering
\caption{Accuracy (\%) on 550 template problems. MCTS-oracle@5 (a search-based upper bound with a perfect verifier) lands \emph{below} BIND for both models, while costing 5--12$\times$ more tokens per problem.}
\label{tab:mcts_oracle}
\footnotesize
\begin{tabular}{lccccc}
\toprule
Model      & Pass@1 & BIND        & Pass@5      & Repair@5 & MCTS-oracle@5 \\
\midrule
GPT-5-Nano & 59.1   & \textbf{82.4} & 82.0       & 78.2     & 78.2 \\
GPT-5      & 86.2   & \textbf{95.8} & \textbf{95.8} & 95.5  & 93.5 \\
\bottomrule
\end{tabular}
\end{table}

\noindent The deployable (no-oracle) SolverLLM variant cannot recognize a correct solution without ground truth, so the search must judge its own output --- a reward signal that is binding-blind. Given its excessive cost, we evaluate it at 10 search iterations on a single small category (disaster\_response, GPT-5-Nano):

\begin{table}[h]
\centering
\caption{Deployable SolverLLM (no oracle) on disaster\_response (GPT-5-Nano). The no-oracle search \emph{explores} a correct solution for 94\% of problems but \emph{returns} a wrong one 30\% of the time, because the self-judge cannot distinguish correct from incorrect solutions. Total cost is $\sim$42$\times$ pass@1 for worse accuracy.}
\label{tab:mcts_no_oracle}
\footnotesize
\begin{tabular}{lccccc}
\toprule
disaster\_response (GPT-5-Nano) & Pass@1 & BIND & Pass@5 & MCTS-oracle@5 & MCTS no-oracle@10 \\
\midrule
Accuracy        & 62\%  & 84\%  & 100\% & 94\%  & \textbf{64\%}  \\
Tokens/problem  & 3.6K  & 1.6K  & 18K   & 27K   & \textbf{152K} \\
\bottomrule
\end{tabular}
\end{table}

\paragraph{Why binding cannot be solved by formulation-level search.} SolverLLM searches over \emph{which formulation} to use, not over \emph{which coefficients} the formulation should bind. On binding-limited categories (which dominate Text2Opt-Bench, see Table~\ref{tab:category_regimes}), the formulation is unambiguous --- the difficulty is faithful transcription of dozens to hundreds of coefficients --- and so re-searching formulations does not help. BIND's data-offloading mechanism removes the transcription burden directly, which is why BIND $\gg$ MCTS-oracle@5 even though MCTS spends an order of magnitude more tokens.

\section{Case Studies: The Impact of BIND}
\label{app:binding_case}

\subsection{Overcoming Binding Failures in Transportation}
We illustrate the binding bottleneck with Qwen2.5-7B on a simple transportation LP (trans\_001). The problem has 7 sources and 6 destinations with supply, demand, and cost data specified in the prompt.

\noindent\textbf{Default (data in prompt) --- FAIL.}
The model correctly identifies the LP structure (continuous variables, supply $\leq$ constraints, demand $=$ constraints, minimize cost) and accurately copies the $7 \times 6$ cost matrix. However, it replaces all supply capacities with a uniform value of 100:

\begin{small}
\begin{verbatim}
# Actual supply: [94, 47, 50, 55, 67, 37, 69]
# Qwen generates:
m.addConstr(quicksum(x[i,j] for j in range(6)) <= 100,
            f"source_capacity_{i}")
# Actual demand: [14, 47, 21, 70, 72, 58]
# Qwen generates:
m.addConstr(quicksum(x[i,j] for i in range(7)) == 100,
            f"destination_requirement_{j}")
\end{verbatim}
\end{small}

This error pattern is common across all 31 transportation failures from Qwen-7B, though the model sometimes gets partially correct numbers.

\noindent\textbf{BIND (data offloaded to file) --- PASS.}
With BIND, numerical data is externalized to a JSON file. The model generates code that \emph{reads} rather than \emph{transcribes} the values:

\begin{small}
\begin{verbatim}
with open(INSTANCE_DATA_PATH, "r") as f:
    d = json.load(f)
# Supply constraint: reads d['supplies'][i] from file
m.addConstr(quicksum(x[i,j] for j in range(d['num_destinations']))
            <= d['supplies'][i], name=f"supply_{i}")
# Demand constraint: reads d['demands'][j] from file
m.addConstr(quicksum(x[i,j] for i in range(d['num_sources']))
            == d['demands'][j], name=f"demand_{j}")
\end{verbatim}
\end{small}

 As we see above, BIND raises Qwen2.5-7B from 38\% to 82\% on transportation ($+$44pp) by eliminating the need to transcribe numerical values, confirming that binding is the bottleneck for this category.

\subsection{BIND Regression on Power Transmission}
\label{app:power_bind_regression}

Power transmission is a modeling-limited category and the only one where BIND regresses for GPT-5 (falling from 98\% to 80\% with refinement, and to 40\% in a single pass). Offloading removes the in-context data scaffold, forcing models to derive physics constraints directly from the schema rather than extracting them from inline values. 

This loss of context disrupts GPT-5's ability to reconstruct unit-conversion chains. Analyzing all BIND-induced failures reveals the dominant error is a spurious conversion factor in the loss coefficient:

\begin{small}
\begin{verbatim}
# WRONG: spurious 1e6 double-counts conversion 
loss_coef = loss_cost_rate * R * (1e6 / (V_kV ** 2))
# CORRECT:
loss_coef = loss_cost_rate * R / (V_kV ** 2)
\end{verbatim}
\end{small}

The model erroneously reasons that power in MW must be multiplied by $10^6$ to reach W, failing to realize that the \texttt{kV} denominator already absorbs this scaling.
\section{Training Details}
\label{app:training-details}

\subsection{Two-Phase Pipeline}

We decompose the optimization solving task into two phases:
\begin{enumerate}
    \item \textbf{Phase 1 (Binding):} A fine-tuned model extracts all decision variables, constraints, and objective function parameters from the natural language problem description into structured JSON.
    \item \textbf{Phase 2 (Solve):} A deterministic template loads the extracted JSON and constructs a Gurobi optimization model programmatically---no LLM is needed.
\end{enumerate}

This decomposition isolates \emph{binding}---the mapping from unstructured text to structured mathematical parameters---as the sole task requiring learned reasoning.

\subsection{Training Data}

We construct binding supervision from the \texttt{resource\_allocation} training split (\texttt{train\_2\_11}), which contains problems with 2--11 decision variables. Each training example pairs a natural language problem description (input) with the corresponding structured JSON extraction (output). The JSON schema includes:
\begin{itemize}
    \item \texttt{goal}: optimization direction (\texttt{MINIMIZE} or \texttt{MAXIMIZE})
    \item \texttt{variables}: list of decision variables with name, type, bounds, and objective coefficient
    \item \texttt{constraints}: list of constraints with coefficients, sense ($\leq$, $\geq$, $=$), and right-hand side
\end{itemize}

The dataset contains 429 examples (387 train / 42 validation after a 90/10 split), with a roughly uniform distribution across variable counts (30--51 examples per variable count from 2 to 11). Average input length is approximately 4{,}200 characters; average output length is approximately 2{,}400 characters.

\subsection{Model Configurations}

We train two binding specialists via full-parameter supervised fine-tuning (SFT) using LLaMA-Factory~\cite{zheng2024llamafactory}:

\begin{table}[h]
\centering
\caption{Binding model training hyperparameters.}
\label{tab:binding-hyperparams}
\begin{tabular}{lcc}
\toprule
\textbf{Hyperparameter} & \textbf{1.5B Binder} & \textbf{7B Binder} \\
\midrule
Base model & Qwen2.5-1.5B-Instruct & Qwen2.5-7B-Instruct \\
Fine-tuning type & Full & Full \\
Epochs & 6 & 6 \\
Learning rate & $1 \times 10^{-5}$ & $1 \times 10^{-5}$ \\
LR scheduler & Cosine & Cosine \\
Warmup ratio & 0.1 & 0.1 \\
Per-device batch size & 2 & 1 \\
Gradient accumulation & 4 & 8 \\
Effective batch size & 8 & 16 \\
Max sequence length & 8{,}192 & 8{,}192 \\
Precision & bf16 & bf16 \\
DeepSpeed & --- & ZeRO Stage 3 \\
Hardware & 1$\times$ A100 80GB & 2$\times$ A100 80GB \\
\bottomrule
\end{tabular}
\end{table}

\subsection{End-to-End SFT Baseline}

For comparison, the end-to-end SFT baseline fine-tunes Qwen2.5-7B-Instruct to directly generate Gurobi solver code from problem descriptions (no intermediate binding step). It is trained on the same variable range (2--11 vars) for 3 epochs with identical learning rate ($1 \times 10^{-5}$), cosine schedule, and DeepSpeed ZeRO-3 configuration.

\subsection{GRPO Training}

We additionally train via Group Relative Policy Optimization (GRPO) using verl by ~\cite{Sheng_2025}. GRPO uses outcome-based rewards from executing generated code against the Gurobi solver, avoiding the need for a learned reward model. Table~\ref{tab:grpo-hyperparams} summarizes the configuration.

\begin{table}[h]
\centering
\caption{GRPO training hyperparameters.}
\label{tab:grpo-hyperparams}
\begin{tabular}{lc}
\toprule
\textbf{Hyperparameter} & \textbf{Value} \\
\midrule
Base model & Qwen2.5-7B-Instruct \\
Algorithm & GRPO \\
Training data & \texttt{train\_2\_11} (vars 2--11) \\
Train batch size & 8 \\
Max prompt length & 4{,}096 \\
Max response length & 4{,}096 \\
Group size ($n$) & 5 \\
Learning rate & $1 \times 10^{-6}$ \\
KL loss & Low-variance KL ($\beta = 0.001$) \\
KL in reward & No \\
Entropy coefficient & 0 \\
Advantage normalization & By std (GRPO default) \\
Rollout engine & vLLM (TP=2) \\
Parallelism strategy & FSDP2 (param + optimizer offload) \\
Total epochs & 15 \\
Save frequency & Every 20 steps \\
Hardware & 4$\times$ A100 80GB \\
Precision & bf16 \\
\bottomrule
\end{tabular}
\end{table}

We experiment with three GRPO variants: (1)~a standard binary reward (1 if the generated code produces the correct optimal objective, 0 otherwise), (2)~an adaptive curriculum sampler that adjusts the sampling distribution across difficulty levels based on an exponential moving average of per-level solve rates ($\alpha = 0.3$, floor weight $= 0.05$), and (3)~a \emph{partial-reward} variant that replaces the sparse binary signal with a hierarchical continuous-credit reward function.

\paragraph{Partial-reward function.}
The partial-reward variant addresses the sparse-reward problem inherent in binary outcome-based RL: most rollouts for hard problems receive zero reward, providing no gradient signal.  We design a hierarchical reward $r \in [0,1]$ that awards incremental credit at successive gates:

\begin{enumerate}
    \item \textbf{Code extraction} (+0.05): valid Python/Gurobi code is parsed from the model output.
    \item \textbf{Execution} (+0.10): the extracted code executes without runtime error.
    \item \textbf{Solver status} (+0.10 if \textsc{optimal}; +0.05 if feasible but not optimal): the Gurobi solver reaches a meaningful termination status.
    \item \textbf{Variable-count match} (+0.05): the number of decision variables in the generated model equals the reference.
    \item \textbf{Constraint satisfaction} (+0.20, continuous): the fraction of reference constraints satisfied by the generated solution, evaluated by substituting generated variable values into the ground-truth constraint matrix.
    \item \textbf{Objective closeness} (+0.20, continuous): $\exp(-\alpha \cdot \text{rel\_gap})$ where $\text{rel\_gap} = |z_{\text{gen}} - z^*| / (|z^*| + 10^{-6})$ and $\alpha = 10$, awarding near-full credit for small deviations and decaying smoothly for larger gaps.
\end{enumerate}

\noindent An exact solution (objective and all variable values within $10^{-4}$ of the reference) overrides the partial score and receives $r = 1.0$.  All other hyperparameters (Table~\ref{tab:grpo-hyperparams}) remain identical across the three GRPO variants.

\paragraph{GRPO results.}
\label{app:grpo}
Table~\ref{tab:grpo-results} compares all GRPO variants against the Qwen2.5-7B-Instruct baseline (zero-shot) on the 248-problem resource allocation eval set.

\begin{table}[h]
\centering
\caption{GRPO variant results on resource allocation (248 eval problems, vars 2--20). In-distribution: vars $\leq 11$; OOD: vars 12--20.}
\label{tab:grpo-results}
\begin{tabular}{lccc}
\toprule
\textbf{Model} & \textbf{Overall} & \textbf{In-dist ($\leq$11)} & \textbf{OOD (12--20)} \\
\midrule
Qwen2.5-7B-Instruct (zero-shot) & 13.3\% (33/248) & 25.0\% (33/132) & 0.0\% (0/116) \\
GRPO (binary reward) & 44.0\% (109/248) & 76.5\% (101/132) & 6.9\% (8/116) \\
GRPO + adaptive curriculum & \textbf{44.8\%} (111/248) & 80.3\% (106/132) & 4.3\% (5/116) \\
GRPO + partial reward & 30.2\% (75/248) & 56.8\% (75/132) & 0.0\% (0/116) \\
\bottomrule
\end{tabular}
\end{table}

All three GRPO variants substantially improve over the zero-shot baseline. The binary-reward and adaptive-curriculum variants perform comparably (44.0\% vs.\ 44.8\%), with the curriculum sampler providing a marginal gain by focusing training on difficulty levels where the model can still learn. The partial-reward variant underperforms at 30.2\%, suggesting that the dense but noisy intermediate credit signal may encourage the model to satisfy partial gates (code extraction, execution, feasibility) without converging to fully correct solutions---a form of reward hacking where the model exploits the hierarchical structure to collect partial credit rather than optimizing for exact correctness. No GRPO variant achieves meaningful OOD generalization beyond vars~11.

\subsection{Evaluation}

All models are evaluated on the full \texttt{eval/} split containing 248 problems with 2--20 decision variables. Problems with 12--20 variables are out-of-distribution (OOD), testing generalization beyond the training range. Inference uses vLLM with greedy decoding (temperature 0, top-$p$ = 1) and a maximum generation length of 4{,}096 tokens.

\subsection{Per-Complexity Breakdown}
\label{app:heatmaps}

\begin{figure}[t]
    \centering
    \includegraphics[width=\textwidth]{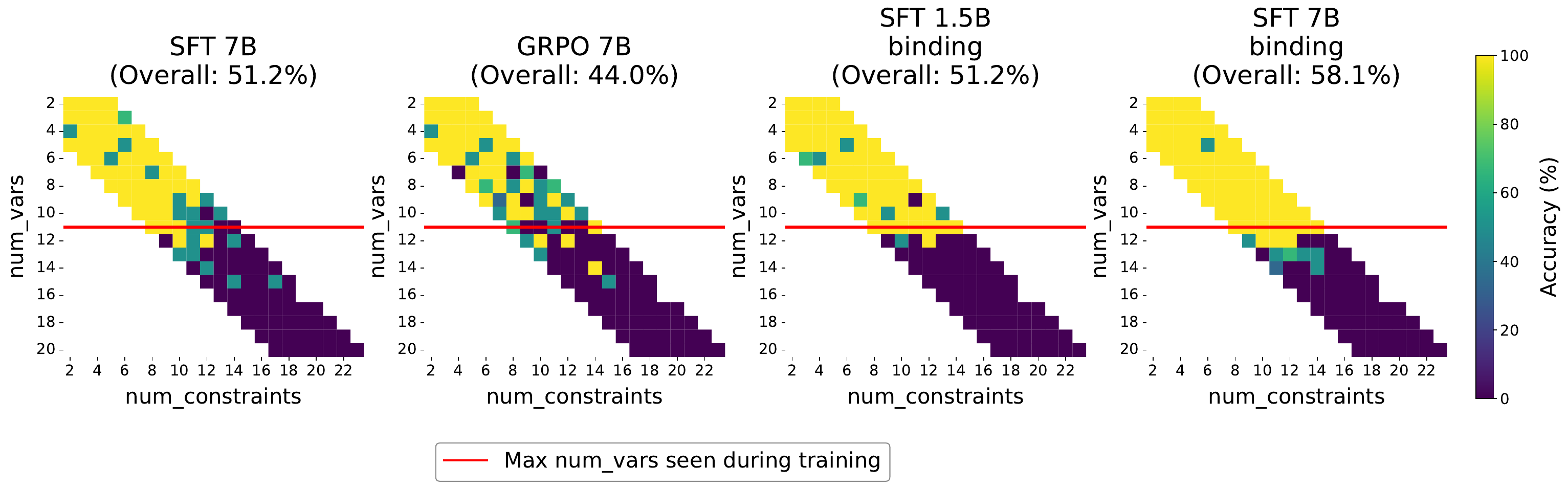}
    \caption{Accuracy heatmaps by problem size (number of variables vs.\
    constraints) for each training approach on resource allocation (248 eval
    problems). The red line marks the maximum number of variables seen during
    training; problems above it are out-of-distribution. Yellow = 100\%
    accuracy, purple = 0\%.}
    \label{fig:heatmaps}
\end{figure}

Figure~\ref{fig:heatmaps} shows accuracy as a function of problem size
(number of variables and constraints) for each training approach. The red
horizontal line marks the maximum number of variables seen during training
($\leq 11$); problems above this line are out-of-distribution (OOD).

The heatmaps reveal several distinct generalization patterns. All approaches learn a sharp in-distribution
boundary: accuracy is near-perfect (yellow) below the red line and collapses
almost entirely (purple) above it, indicating that none of the training regimes
generalize binding to larger problem sizes. The 7B binding specialist
shows the cleanest in-distribution coverage, while SFT 7B end-to-end exhibits scattered failures
even on seen problem sizes, consistent with imperfect
binding under joint training. GRPO shows the most irregular
in-distribution pattern, with high variance across cells of similar complexity,
reflecting the difficulty of learning precise coefficient transcription from a
sparse, binary reward.

\subsection{Multi-Category Binding Specialists}
\label{app:multi-category-binding}

We extend the binding hypothesis to two additional OR problem categories: transportation
(LP) and JSSP (MILP). Both use Qwen2.5-7B-Instruct with the same training configuration
as the resource allocation binder (Table~\ref{tab:binding-hyperparams}): full-parameter SFT, 6 epochs, lr=$1\times10^{-5}$, cosine
schedule, ZeRO-3 on 2$\times$A100 80GB\@. Phase~2 uses deterministic template code that loads
the extracted JSON and constructs a Gurobi model programmatically.

\paragraph{Training data.}
Transportation: 244 instances with sources $\times$ destinations $\leq 52$.
JSSP: 224 instances with $n_\text{jobs} \leq 5$.
Both drawn from the respective \texttt{Template\_train/} splits.

\paragraph{Evaluation.}
50 problems per category. Transportation: 26 in-distribution
(sources $\times$ destinations $\leq 52$), 24 OOD (up to $25 \times 25 = 625$ variables).
JSSP: 25 in-distribution (jobs $\leq 5$), 25 OOD (jobs 6--13, up to 52 operations).
The GT$\to$template upper bound achieves 100\% on both categories, confirming that binding is the sole bottleneck.

\paragraph{JSSP results.}
The binding specialist achieves 100\% overall (50/50),
including 100\% OOD (25/25). End-to-end SFT achieves 96.0\% (48/50), with 2 OOD failures
from code generation errors.

\paragraph{Transportation results.}
The binding specialist achieves 96.0\% overall (48/50) vs.\ 88.0\% for end-to-end SFT
(44/50), with OOD accuracy of 91.7\% vs.\ 75.0\%. A key implementation detail: the binding target uses \emph{compact} JSON (no indentation, \texttt{separators=(',',':')}), which reduces output token length by ${\sim}3\times$ compared to indented JSON for large cost matrices. In an initial experiment using indented JSON, the binding specialist achieved only 80.0\% vs.\ 90.0\% for end-to-end SFT---the token-length disadvantage caused attention copy errors on OOD instances.

\paragraph{Description format robustness.}
We additionally evaluate both approaches with prose descriptions (per-source sentences
with randomized destination ordering, no tables). Both models degrade proportionally:
JSSP drops ${\sim}20$pp OOD for both (binding: 60\%, end-to-end: 64\%); transportation
drops similarly. The parallel degradation confirms that the binding bottleneck is output
sequence length, not input description complexity.

\subsection{OOD Cliff-Shift Experiment}
\label{app:ood-shift}

To test whether the OOD gap reflects limited training
coverage or a fundamental extraction limit, we train a second 7B binding
specialist on vars 2--15 (565 examples, same configuration as Table~\ref{tab:binding-hyperparams})
and evaluate on the same 248-problem eval set.

\begin{table}[h]
\centering
\scriptsize
\caption{Effect of training coverage on binding specialist accuracy
(resource allocation, 248 eval problems). The OOD cliff shifts from
var=11 to var=15, with no regression on the original in-distribution range.}
\label{tab:ood-shift}
\begin{tabular}{l ccc c}
\toprule
\textbf{Training range} & \textbf{Model} & \textbf{Overall} & \textbf{In-dist} & \textbf{OOD} \\
\midrule
vars 2--11 & 7B binding specialist & 58.1\% (144/248) & 99.2\% (131/132, $\leq$11) & 11.2\% (13/116, 12--20) \\
vars 2--11 & 7B end-to-end SFT & 51.2\% (127/248) & 88.6\% (117/132, $\leq$11) & 8.6\% (10/116, 12--20) \\
\midrule
vars 2--15 & 7B binding specialist & \textbf{75.4\%} (187/248) & 91.8\% (169/184, $\leq$15) & 28.1\% (18/64, 16--20) \\
vars 2--15 & 7B end-to-end SFT & 57.3\% (142/248) & 75.5\% (139/184, $\leq$15) & 4.7\% (3/64, 16--20) \\
\midrule
Any & Ground truth $\to$ template & 100\% & 100\% & 100\% \\
\bottomrule
\end{tabular}
\end{table}

Three findings emerge. First, the cliff shifts: overall accuracy improves from
58.1\% to 75.4\%, and the binding specialist's advantage over end-to-end SFT
widens from +6.9pp to +18.1pp. Second, accuracy on vars 2--11 is preserved,
ruling out catastrophic forgetting. Third, vars 12--15 reach 80.8\%---below the
99.2\% achieved by the original model on vars 2--11---reflecting the genuine
difficulty of extracting 50+ coefficients from longer prose, not a training artifact.
These results confirm that the OOD gap reflects training coverage, not a
fundamental limit of binding SFT.

\subsection{Additional RL Ablations: SFT-Warm-Started GRPO and Curriculum GRPO}
\label{app:rl_ablations}

We ran two additional RL ablations on resource allocation to test whether stronger
initial conditions or a difficulty curriculum let RL catch up to SFT for binding-heavy
tasks. Neither does.

\paragraph{Warm-starting GRPO from SFT checkpoints (Table~\ref{tab:rl_warmstart}).}
RL improves overall accuracy on only 3 of 8 checkpoints, marginally and only for
under-trained ones. It degrades the best-converged checkpoints and collapses the
weakest seed. No checkpoint beats dense SFT (51.2\%) or cold-start GRPO (44.0\%).

\begin{table}[h]
\centering
\caption{Warm-starting GRPO from SFT checkpoints (resource allocation). $\Delta$ in pp.}
\label{tab:rl_warmstart}
\footnotesize
\begin{tabular}{rrrr}
\toprule
SFT ckpt (step) & SFT overall & +RL overall & $\Delta$ \\
\midrule
20  & 7.7\%  & 0.0\%  & $-7.7$ \\
40  & 17.3\% & 22.2\% & $+4.8$ \\
60  & 29.8\% & 29.8\% & $\phantom{-}0.0$ \\
80  & 31.9\% & 33.5\% & $+1.6$ \\
100 & 35.1\% & 34.3\% & $-0.8$ \\
120 & 35.1\% & 35.5\% & $+0.4$ \\
140 & 37.9\% & 35.1\% & $-2.8$ \\
153 & 37.5\% & 34.7\% & $-2.8$ \\
\bottomrule
\end{tabular}
\end{table}

\paragraph{Curriculum (same seed + difficulty schedule, Table~\ref{tab:rl_curriculum}).}
The model is first warm-started via SFT on easy problems ($\le 2$ variables) and then trained
on progressively harder ones. We compare continuing the curriculum with RL (GRPO) versus
SFT, from the same seed and data. While the model improves via RL over the course of the
curriculum, it falls behind SFT from stage 5 onward (28.6\% vs 37.5\% at the final stage;
51.2\% for full SFT). This corroborates that SFT is better suited to binding-heavy
optimization tasks.

\begin{table}[h]
\centering
\caption{Curriculum learning: GRPO vs SFT from the same seed and difficulty schedule.}
\label{tab:rl_curriculum}
\footnotesize
\begin{tabular}{lrrr}
\toprule
Difficulty (\# variables) & GRPO & SFT & $\Delta$ (GRPO$-$SFT) \\
\midrule
seed ($\le 2$) & 8.9\%  & 8.9\%  & $\phantom{-}0.0$ \\
$\le 3$  & 12.9\% & 11.7\% & $+1.2$ \\
$\le 4$  & 19.8\% & 19.8\% & $\phantom{-}0.0$ \\
$\le 5$  & 19.8\% & 17.7\% & $+2.0$ \\
$\le 6$  & 20.6\% & 25.4\% & $-4.8$ \\
$\le 7$  & 21.8\% & 28.2\% & $-6.5$ \\
$\le 8$  & 20.6\% & 32.7\% & $-12.1$ \\
$\le 9$  & 27.4\% & 31.5\% & $-4.0$ \\
$\le 10$ & 28.6\% & 37.5\% & $-8.9$ \\
\bottomrule
\end{tabular}
\end{table}

\paragraph{Reward shaping note.} Our partial reward function is already milestone-gated:
no reward until parseable code, then execution, then valid solver status. A middle ground
(reward only for parseable+executable+feasible code) is plausible but not pursued, given
SFT's relative strength and RL's cost.

\subsection{Binding Specialist vs End-to-End SFT: Paired McNemar}
\label{app:specialist_vs_e2e}

On the same evaluation sets, the binding specialist (trained on binding only) is compared head-to-head against an end-to-end SFT model of identical scale (Qwen2.5-7B, same data, same recipe modulo target). The discordant counts $b/c$ are the per-problem wins of specialist vs E2E SFT.

\begin{table}[h]
\centering
\caption{Binding specialist vs end-to-end SFT (paired McNemar). On resource allocation the specialist advantage is statistically significant (p=0.002), driven by a large in-distribution gain. On the two fixed-schema categories the specialist is directionally better and never worse (small $n$ limits power).}
\label{tab:specialist_vs_e2e}
\footnotesize
\begin{tabular}{lcccccc}
\toprule
Category & n & Specialist (95\% CI) & E2E SFT (95\% CI) & $\Delta$ & b/c & McNemar p \\
\midrule
Resource allocation & 248 & 58.1\% [52, 64] & 51.2\% [45, 57] & +6.9 & 23/6 & \textbf{0.002} \\
JSSP                & 50  & 100.0\% [93, 100] & 96.0\% [87, 99] & +4.0 & 2/0  & 0.500 \\
Transportation      & 50  & 96.0\% [87, 99]  & 88.0\% [76, 94] & +8.0 & 4/0  & 0.125 \\
\bottomrule
\end{tabular}
\end{table}

\subsection{Multi-Task Binding Specialist}
\label{app:multitask_specialist}

We train a single Qwen2.5-7B binding specialist jointly on a mixture of resource allocation, transportation, and JSSP. We use the same hyperparameters as the per-category specialists (effective batch 16, 3 epochs, LR $10^{-5}$, BF16). We compare this multi-task binder against both its per-category counterparts (one model per task) and a multi-task end-to-end SFT baseline (trained on the same mixture to emit solver code directly instead of structured JSON).

\begin{table}[h]
\centering
\caption{Multi-task binding specialist vs.\ multi-task end-to-end SFT (Qwen2.5-7B). On resource allocation, the multi-task binder significantly outperforms the end-to-end model (60.5\% vs 45.6\%) and slightly exceeds the per-category binder (58.1\%). On JSSP and transportation, it remains competitive. While a multi-task binder is viable, it is not uniformly superior; scaling to a universal binder is left to future work. Sample sizes: RA n=248 (132 ID / 116 OOD); Transportation n=50 (26 ID / 24 OOD); JSSP n=50 (25 ID / 25 OOD). Wilson 95\% CIs in brackets.}
\label{tab:multitask}
\footnotesize
\setlength{\tabcolsep}{4pt}
\resizebox{\textwidth}{!}{%
\begin{tabular}{lcccccc}
\toprule
Task & \multicolumn{2}{c}{Overall} & \multicolumn{2}{c}{In-distribution} & \multicolumn{2}{c}{OOD} \\
\cmidrule(lr){2-3} \cmidrule(lr){4-5} \cmidrule(lr){6-7}
 & MT Binder & MT End-to-end & MT Binder & MT End-to-end & MT Binder & MT End-to-end \\
\midrule
Resource allocation & \textbf{60.5} [54, 66]   & 45.6 [39, 52]            & \textbf{99.2} [96, 100]  & 85.6 [79, 91]            & \textbf{16.4} [11, 24]   & 0.0 [0, 3] \\
Transportation      & 84.0 [71, 92]            & \textbf{90.0} [79, 96]   & \textbf{100.0} [87, 100] & \textbf{100.0} [87, 100] & 66.7 [47, 82]            & \textbf{79.2} [60, 91] \\
JSSP                & \textbf{100.0} [93, 100] & \textbf{100.0} [93, 100] & \textbf{100.0} [87, 100] & \textbf{100.0} [87, 100] & \textbf{100.0} [87, 100] & \textbf{100.0} [87, 100] \\
\bottomrule
\end{tabular}}
\end{table}

\end{document}